\documentclass[twocolumn]{article}
\usepackage[margin=0.75in]{geometry}
\usepackage{cite}
\usepackage{amsmath,amssymb,amsfonts}
\usepackage{algorithmic}
\usepackage{graphicx}
\usepackage{textcomp}
\usepackage{bmpsize}
\usepackage{xcolor}
\usepackage{lipsum}
\usepackage{hyperref}
\usepackage{subfig} 
\usepackage{listings}
\usepackage{xcolor}
\usepackage{booktabs}
\usepackage{bm}
\usepackage{amsthm}
\usepackage{titling}
\usepackage{threeparttable}
\usepackage{multirow}
\usepackage{makecell}
\usepackage{subcaption}

\theoremstyle{definition}
\newtheorem{definition}{Definition}[section] 

\definecolor{purple}{HTML}{9467BD} 
\definecolor{olive}{HTML}{BCBD22} 
\definecolor{cyan}{HTML}{17BECF} 

\pretitle{\centering\Large\bfseries}
\posttitle{\par\vspace{1em}}

\preauthor{\centering\normalsize\lineskip 0.5em}
\postauthor{\par}

\predate{\centering\small}
\postdate{\par\vspace{1.5em}}

\begin{document}

\title{Free Record-Level Privacy Risk Evaluation Through Artifact-Based Methods}

\date{}

\author{
Joseph Pollock\footnotemark[1] \quad 
Igor Shilov\footnotemark[1] \quad 
Euodia Dodd \quad 
Yves-Alexandre de Montjoye
\\[1ex]
\textit{Imperial College London}
}

\maketitle

\def\thefootnote{*}\footnotetext{Equal contribution}\def\thefootnote{\arabic{footnote}}

\begin{abstract}
Membership inference attacks (MIAs) are widely used to empirically assess privacy risks in machine learning models, both providing model-level vulnerability metrics and identifying the most vulnerable training samples. State-of-the-art methods, however, require training hundreds of shadow models with the same architecture as the target model. This makes the computational cost of assessing the privacy of models prohibitive for many practical applications, particularly when used iteratively as part of the model development process and for large models.
We propose a novel approach for identifying the training samples most vulnerable to membership inference attacks by analyzing artifacts naturally available during the training process. Our method, Loss Trace Interquartile Range (LT-IQR), analyzes per-sample loss trajectories collected during model training to identify high-risk samples without requiring any additional model training.
Through experiments on standard benchmarks, we demonstrate that LT-IQR achieves 92\% precision@k=$1\%$ in identifying the samples most vulnerable to state-of-the-art MIAs. This result holds across datasets and model architectures with LT-IQR outperforming both traditional vulnerability metrics, such as loss, and lightweight MIAs using few shadow models. We also show LT-IQR to accurately identify points vulnerable to multiple MIA methods and perform ablation studies. We believe LT-IQR enables model developers to identify vulnerable training samples, for free, as part of the model development process. Our results emphasize the potential of artifact-based methods to efficiently evaluate privacy risks.
\end{abstract}

\section{Introduction}
\label{sec:introduction}
\begin{figure}
    \centering
    \includegraphics[width=0.4\textwidth]{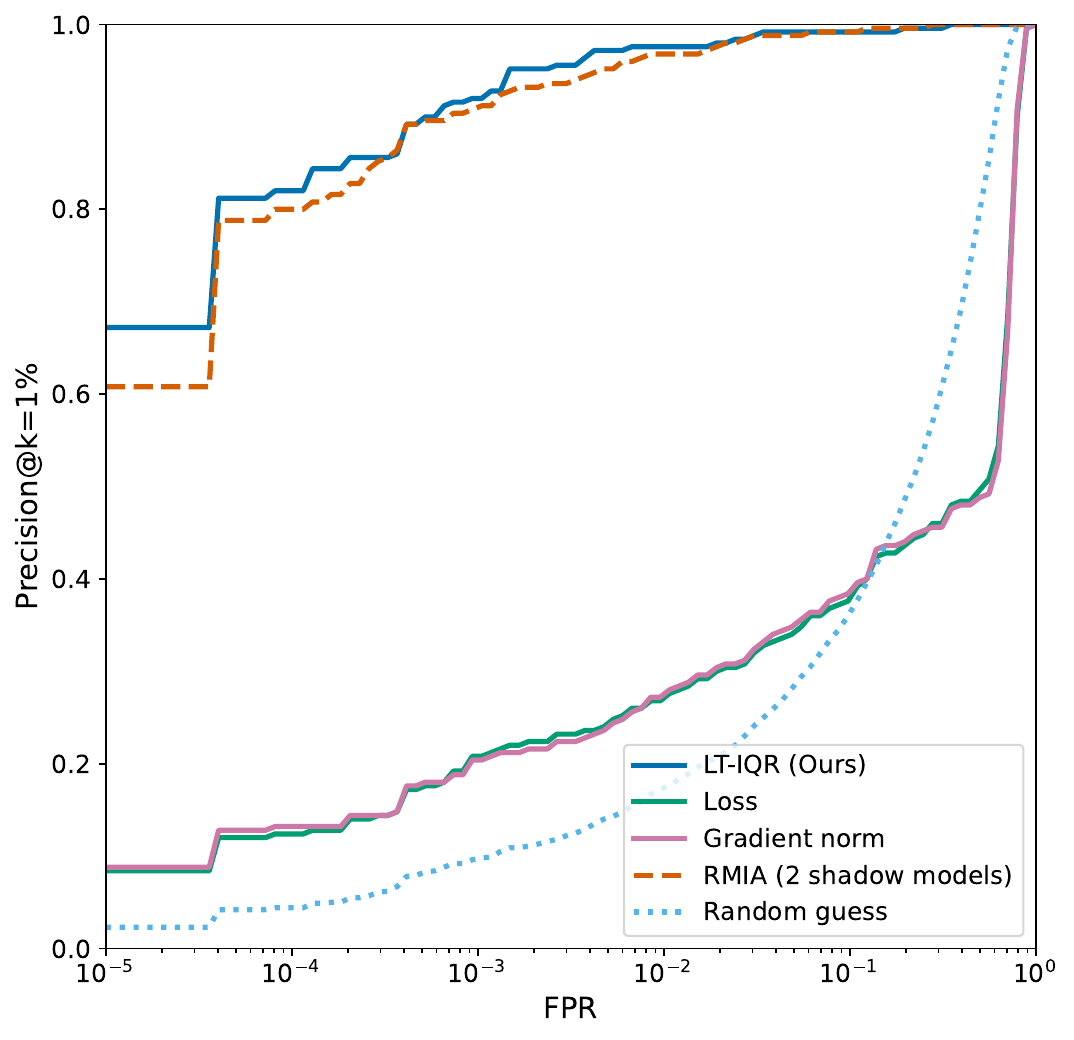}
    \caption{Precision@k=1\% (250 samples) when identifying vulnerable samples determined by LiRA attack at a variable FPR threshold}
    \label{fig:precision_at_1}
\end{figure}

In recent years, machine learning (ML) models have become increasingly prevalent within sensitive domains. Companies, governments, and academic researcher are now routinely training and fine-tuning these models using confidential or sensitive data from areas such as the healthcare~\cite{kaissis2020secure}, legal~\cite{cui2023chatlaw}, and financial~\cite{asif2024us} sectors.

ML models have been shown to memorize their training data~\cite{song2017machine,carlini2023quantifying,carlini2022membership,meeus2024copyright}, potentially allowing an adversary to detect the presence of a certain data sample in the training dataset~\cite{yeom2018privacy,shokri2017membership,carlini2022membership} and, in some cases, to even reconstruct training samples~\cite{balle2022reconstructing,fredrikson2015model}.

Membership inference attacks (MIAs) are widely used by model developers to empirically evaluate the privacy risk associated with the release of the model~\cite{yeom2018privacy, carlini2022membership, ye2022enhanced}. Running an MIA against their own model allows developers to learn two valuable insights. First, it provides a sense of the model-level risk, the number of points an adversary might be able to confidently identify as members of the training dataset typically measured as True Positive Rate (TPR) at a certain False Positive Rate (FPR) threshold. Second, it identifies the points that are most likely to be vulnerable to an attack - e.g. the training data points where the attack confidence is above the threshold for a given FPR.

Model developers will typically evaluate the overall privacy risk by taking into consideration the estimated model-level risk (e.g. TPR@low FPR), the sensitivity of the data, the anticipated threat model (white-box or black-box) and model deployment strategy, as well as the analysis of the specific points that might be a risk (e.g. to ensure there is no disparate impact). Based on these, the developer would determine if releasing the model is deemed safe from a privacy perspective. As such, vulnerability to MIAs is an important factor in determining whether a model is considered anonymous including from a legal perspective as emphasized in the opinion published by the European Data Protection Board on AI last month~\cite{edpb_opinion_28_2024}.

When the risk of the utility-maximizing model--the model developed without taking privacy considerations into account--is deemed too high, defenses will be put in place. This includes both privacy-preserving training methods using formal guarantees, such as DP-SGD~\cite{abadi2016deep}, as well as approaches--including approaches targeted at specific points--such as increased regularization, reducing the model size, removing the most vulnerable points from the training data (which has been showed to decrease the overall risk~\cite{carlini2022privacy}) or unlearning them from the trained model. Knowing both which and how many points are vulnerable to an MIA is essential to the development and evaluation of defense strategies.

State-of-the-art privacy attacks are, however, computationally expensive to run. They require training multiple, often hundreds, of shadow models~\cite{carlini2022membership,ye2022enhanced,zarifzadeh2024low}, with each shadow model requiring the same amount of computational resources as one original training run. Their computational complexity increases with model and dataset scale, making such attacks impractical if not infeasible—particularly for large language models—and significantly constraining their utility in processes like architecture or training parameter selection.

\textbf{Contribution.} We here propose a new class of artifact-based methods to estimate privacy risks which we instantiate on the task of detecting points that might be vulnerable to MIAs. Our method leverages access to training artifacts to detect the training samples most at risk of being identified as members by state-of-the-art MIAs. We define a training data point as vulnerable if a given MIA confidently and correctly predicts its membership, i.e. the point is predicted as a member at a fixed low False Positive Rate (FPR). 

As our main contribution, we propose Loss Trace Interquartile Range (LT-IQR), a novel method analyzing the per-sample loss dynamics throughout training to identify the most vulnerable training samples. It is based on the intuition that points prone to be memorized are likely to be outliers and therefore harder for the model to learn. This would lead to the loss of these samples converging slower than that of an average point. Specifically, LT-IQR computes the interquartile range across intermediate losses collected once per epoch, and is shown to be a good predictor of individual vulnerability. 

LT-IQR achieves 92\% Precision@k=1\% in predicting training samples from CIFAR-10~\cite{Krizhevsky09learningmultiple} vulnerable to Likelihood Ratio Attack (LiRA)~\cite{carlini2022membership} at a low FPR threshold of $10^{-3}$, significantly outperforming traditional vulnerability predictors based on metrics like final loss or gradient norms. This advantage holds across datasets (CIFAR-10, CIFAR-100, and CINIC-10) and model architectures (ResNet-20, WideResNet28-2, WideResNet40-4). Notably, LT-IQR even outperforms lightweight membership inference attacks that require training shadow models (like RMIA~\cite{zarifzadeh2024low} with 2 shadow models) when we use the same Precision@k metric to measure the agreement between the attacks.

While initially focused on predicting samples vulnerable specifically to the LiRA attack, we later expand the definition and focus on more generic vulnerability to \textit{any} of the widely adopted state-of-the-art MIAs: LiRA~\cite{carlini2022membership}, Attack R~\cite{ye2022enhanced} and RMIA~\cite{zarifzadeh2024low}. Impressively, when measured on predicting vulnerability to at least one of these attacks, our method performs on par with Attack R and RMIA with a few shadow models, despite incurring a fraction of the computational cost.

We then explore other loss trace aggregation methods - all of which analyze the per-sample training dynamics captured by the per-epoch loss trace. We find that multiple aggregation methods show strong performance on par with LT-IQR, with slight variability depending on the dataset. Additionally, we perform an ablation study on LT-IQR's quantile parameters, showing the method to be quite robust to specific parameter values.

Finally, we analyze the distribution of LT-IQR vulnerability scores throughout training, and note the distinctive changes in the distribution shape as the training progresses and the model memorizes more. This suggests that LT-IQR could also potentially be used to estimate model-level vulnerability in addition to detecting vulnerable points, as considered before.

Our cheap yet effective approach enables model developers to assess the privacy risks associated with a given model release and deploy appropriate mitigations if needed, including unlearning, data removal, or retraining with differential privacy. Given its low computational cost, our method could be especially useful to compare the privacy leakage of different hyperparameter setups, or as an early indication to run a full privacy attack on a subset of candidate models.

\section{Related Work}
\label{sec:related_work}
\subsection{Membership Inference Attacks}
\label{sec:related_work:mias}

Membership Inference Attacks (MIAs) aim to determine whether a certain data sample was used to train a given model. They typically rely on a strong yet realistic attacker~\cite{ponomareva2023dp,murakonda2020ml}, and are used by model developers as an upper bound for other attacks including extraction ~\cite{carlini2021extracting,nasr2021adversary,lukas2023analyzing} and attribute inference. In many cases the vulnerability to an MIA does not pose a high risk in practice: either because a realistic attacker would not have access to the full user record, or because membership alone does not imply the possession of a sensitive attribute. However, as MIA is the easiest type of attack for an adversary to perform, it is often used to estimate empirical privacy risk, under the assumption that protection against an MIA implies the protection against all other types of attack~\cite{ponomareva2023dp}.

MIAs are typically evaluated with TPR at low FPR - a metric measuring an attacker's ability to confidently and correctly predict membership of a small number of points, irrespective of the attack's performance on average. This recognizes the fact that in a privacy setup, an attacker making confident predictions about a small minority of points possesses higher risk than an attacker making predictions slightly better than random on every point.

Widely adopted MIAs today, including LiRA~\cite{carlini2022membership} and Attack R~\cite{ye2022enhanced}, however involve training a large number of \textit{shadow models}~\cite{shokri2017membership,carlini2022membership,long2020pragmatic,ye2022enhanced}. This makes their practical application very expensive, and in many cases infeasible. 

Existing low-cost attacks~\cite{yeom2018privacy,zarifzadeh2024low} either compromise utility relative to shadow-model-based approaches or still necessitate training some shadow models~\cite{zarifzadeh2024low}, which remains computationally expensive given the positive correlation between attack performance and the number of shadow models employed.

\subsection{Individual vulnerability}
\label{sec:related_work:individual_vulnerability}

An essential step in DP-SGD~\cite{abadi2016deep} (a differentially private variation of SGD) is clipping the per sample gradient to bound the sensitivity of the output to any individual sample. This demonstrates the disparate nature of individual data points' contribution to the model - where clipping is necessary to ensure the model doesn't learn too much from any one sample.

In the context of differentially private training~\cite{dwork2006calibrating, abadi2016deep}, individual data-dependent privacy accounting has been proposed, recognizing non-uniform privacy leakage across training data samples~\cite{yu2023individual,koskela2023individual,thudi2024gradients}. Unlike classic data-independent privacy accounting mechanisms~\cite{abadi2016deep,mironov2017renyi,mironov2019r}, individual privacy accounting considers per-sample gradient norms and yields stronger privacy guarantees for data points falling below the clipping threshold. As such, Thudi et al.~\cite{thudi2024gradients} show that on common benchmark datasets, many individual data samples enjoy significantly better privacy protection than what is suggested by the conservative data-independent privacy analysis traditionally used for DP-SGD.

It then naturally follows, that in the absence of gradient clipping, some points would be memorized more than others. Empirical privacy risk is indeed typically concentrated around a small number of data outliers~\cite{feldman2020neural,bagdasaryan2019differential,carlini2022privacy}. 

Specifically from the perspective of MIA vulnerability, two prior works explore the correlation between the attack score and certain data properties. Leemann et al.~\cite{leemann2024my} investigate the connection between MIA score and certain per-instance properties: final loss, parameter gradients, local loss Hessian, and others, identifying the final loss to be the best predictor of the MIA performance. They conclude that the loss is the best predictor of the MIA vulnerability. Tobaben et al.~\cite{tobaben2024impactdatasetpropertiesmembership} study the impact of the average gradient norm throughout training on the MIA performance and find a strong correlation across multiple datasets. Unlike our work, however, both Leemann et al.~\cite{leemann2024my} and Tobaben et al.~\cite{tobaben2024impactdatasetpropertiesmembership} study the correlation between the data properties and attack success while we focus on the few most vulnerable samples and using training artifacts as a source of information.

\subsection{Tracing memorization}
\label{sec:related_work:tracing_memorization}

Unlike privacy analyses described above, we consider an artifact-based approach where the model developer has access to all intermediate model states during training. This setup allows us to use the loss throughout training, for instance, as an additional source of information on top of the commonly used final loss. 

Some prior works have considered tracing privacy metrics throughout training. Pruthi et al.~\cite{pruthi2020estimating} tracks the loss before and after each training step to estimate the \textit{influence} of a training sample on model's predictions. Lesci et al.~\cite{lesci2024causal} characterize a model's counterfactual memorization~\cite{zhang2023counterfactual} by observing losses on a subset of data throughout training. Liu et al.~\cite{liu2022membership} and Li et al.~\cite{li2024seqmiasequentialmetricbasedmembership} propose an MIA where an attacker performs model distillation and tracks the loss throughout the distillation process.

For differentially private training, \textit{privacy auditing} often adopts a similar threat model to us. Privacy auditing has emerged as a way to either assess the tightness of the theoretical privacy analysis~\cite{jagielski2020auditing, nasr2021adversary, zanella2023bayesian} or verify the correctness of DP algorithm implementations~\cite{tramer2022debugging,nasr2023tight}. Privacy auditing empirically estimates the lower bound for the differential privacy guarantees of the model training process by running privacy attacks and measuring their success. Assumed threat models can range from a weak adversary with only black-box access to the model~\cite{jagielski2020auditing,nasr2021adversary,steinke2024privacy}, to a white-box adversary capable of observing intermediate model updates~\cite{nasr2021adversary,nasr2023tight,steinke2024privacy}, to a very strong adversary with full control over the training dataset apart from the target record~\cite{nasr2021adversary}. 

Proposed attack methods vary depending on the threat model. For black-box attacks, a simple loss threshold~\cite{yeom2018privacy} is typically used. For white-box adversaries with access to the intermediate model states, a threat model similar to the one we assume in this work, prior works suggested using aggregate metrics over the entire training process. Proposed methods include taking the maximum or average of the loss~\cite{nasr2021adversary}, or computing the dot product between the canary gradient and the overall batch update~\cite{nasr2023tight, steinke2024privacy}. 

As the goal of the privacy auditing is to provide a tight estimate of the differential privacy guarantees, prior works typically rely on data poisoning attacks, instantiating the MIAs against data samples (\textit{canaries}) purposely injected into the training dataset. This allows the auditor to construct the worst possible scenario with the maximum privacy leakage. In contrast, we here focus on estimating the privacy risks of realistic data points in a non-adversarially constructed dataset.

\section{Task}
\label{sec:task}

In this paper we set the task of identifying the most vulnerable points among the training samples. Given the training dataset, we aim to rank the data points by their vulnerability, providing a ranked list as an output.
As the privacy risk is typically concentrated in a small proportion of the training dataset - and only this subset is highly relevant for any potential mitigation - we focus our evaluation on the top end of the output ranked list.

For evaluation purposes we assume access to binary ground truth vulnerability labels representing actual attack results (which we expect to be highly imbalanced). Given these labels, we evaluate our vulnerability detection methods using Precision@k and Recall@k, which measure what fraction of the k points with highest vulnerability scores are actually vulnerable. For a vulnerability scoring function $\phi$ and a set of vulnerable points $V$, we define Precision@k as:

\[
\text{Precision@k} = \frac{|\text{top}_k(\phi) \cap V|}{k}
\]

where $\text{top}_k(\phi)$ represents the k points with the highest vulnerability scores according to $\phi$. In other words, if we sort all training points by their vulnerability score $\phi(x,y)$ in descending order and take the first k points, Precision@k tells us what fraction of these points are actually vulnerable.

Similarly, we define Recall@k as:

\[
\text{Recall@k} = \frac{|\text{top}_k(\phi) \cap V|}{|V|}
\]

Here, recall measures the fraction of all vulnerable points we successfully identified within our top k predictions.

For acquiring the ground truth labels, we perform one or multiple state-of-the-art shadow model-based MIAs, characterized by their ROC curves. We consider a fixed and low FPR threshold $\alpha$ (typically $\alpha = 10^{-3}$) and define samples that are identified as members at this strict threshold as vulnerable. We use the terms ``point'' and ``sample'' interchangeably. For a given $\alpha$, we select the threshold $\tau_\alpha$ such that $\text{FPR}(\tau_\alpha) = \alpha$.

\begin{definition}[Vulnerable points]
\label{def:vulnerable}
We define the set of vulnerable points $V$ at a given FPR threshold $\alpha$ as:

\begin{equation}
    V_\alpha = \{(x,y) \in D_{\text{train}} : A(x,y,\theta) \geq \tau_\alpha\}
\end{equation}
\end{definition}

This allows us to evaluate the precision of our vulnerability detection method by comparing against the set of vulnerable points identified by SOTA shadow model-based (SM-based) MIA techniques. The focus on vulnerable samples identified at a strict FPR threshold $\alpha$ captures the samples that are most clearly distinguishable as training set members. These points represent the highest privacy risk, as they can be confidently identified by SOTA SM-based MIAs as members even under conservative attack settings that minimize false positives. By fixing a low FPR threshold (e.g., $\alpha = 10^{-3}$), we ensure that any point classified as vulnerable represents a genuine privacy concern rather than a statistical artifact, since the attacker must be highly selective to maintain such a low false positive rate. 

We initially consider vulnerability to LiRA, as it is the best performing attack with the highest number of vulnerable points - a clear choice for a hypothetical attacker. Additionally, we show that most points vulnerable to other attacks are also vulnerable to LiRA, but not the other way around. Only about 7\% of points vulnerable to RMIA and Attack R are not covered by LiRA (see Fig.~\ref{fig:venn}).

We also note that perfect Recall@k is often not achievable due to the relationship between k and the total number of vulnerable points identified at a given FPR threshold. For instance, with $k=1\%$ (250 samples) and 2435 total vulnerable points in our standard setup, the maximum theoretical recall is only 10\%. We report the maximum achievable Recall@k values for each experimental configuration and discuss this property further in Appendix~\ref{app:recall}.

\section{Preliminaries}
\label{sec:preliminaries}
\subsection{Problem setup}
\label{sec:preliminaries:problem_setup}

\subsubsection{Model training}
\label{sec:preliminaries:problem_setup:model_training}

Let $f_\theta: \mathcal{X} \rightarrow [0,1]^N$ be a machine learning classification model with an input domain $\mathcal{X}$ and an output domain $\mathcal{Y}$ with $N=|\mathcal{Y}|$ classes, parametrized by a parameter vector $\theta$. $f_\theta(x)$ outputs a probability distribution over $N$ classes. For brevity, we denote the probability of a given class $y$ as $f_\theta(x)_y$. We denote train and test datasets as $D_\text{train} = \{(x_i, y_i) \mid x_i \in \mathcal{X}, y_i \in \mathcal{Y}\}$ and $D_\text{test} = \{(x^*_i, y^*_i) \mid x^*_i \in \mathcal{X}, y^*_i \in \mathcal{Y}\}$, respectively. Let $\mathcal{T}$ be a training process such that the final model weights $\theta$ are obtained as $\theta \leftarrow \mathcal{T}(D_\text{train})$. We consider the training process to be performed over $S$ epochs, and additionally denote the intermediate model state after $s = 1 \dots S$ epochs as $\theta_s \leftarrow \mathcal{T}_s(D_\text{train})$. Note that $\theta_{0}$ is the initial model state and $\theta_S = \theta$ is the parameter vector of a fully trained model. 

We consider training with cross-entropy loss defined as:

\begin{equation}
\ell(f_\theta(x),y) = - \log{f_\theta(x)_y}
\end{equation}

Specifically for our method, we denote the loss on a sample $(x,y)$ after $s$ training epochs as $\ell_s(x,y) = \ell(f_{\theta_s}(x),y)$. We also define $\mathcal{L}_\text{train}^s$ and $\mathcal{L}_\text{test}^s$ as average losses at epoch $s$ on train and test sets respectively.

\begin{equation}
\mathcal{L}_{\text{train}}^s = \frac{1}{|D_{\text{train}}|} \sum_{(x,y) \in D_{\text{train}}} \ell_s(x,y)
\end{equation}

\begin{equation}
\mathcal{L}_{\text{test}}^s = \frac{1}{|D_{\text{test}}|} \sum_{(x^*,y^*) \in D_{\text{test}}} \ell_s(x^*,y^*)
\end{equation}

Training is performed in mini-batches $\mathcal{B}_i$ with each optimization step defined as

\begin{equation}
\theta^{t+1} \leftarrow \theta^t - \eta \sum_{(x,y) \in \mathcal{B}_i} \nabla_\theta \ell(f_\theta(x), y)
\end{equation}

where $\Tilde{\theta}$ is the updated parameter vector and $\eta$ is the learning rate.

\subsection{Membership Inference Attacks}
\label{sec:preliminaries:problem_setup:mias}

A membership inference attack (MIA) aims to determine whether a given data sample was used to train a model. Formally, we define a membership inference attack as a function $A: \mathcal{X} \times \mathcal{Y} \times \Theta \rightarrow [0,1]$, where $A(x, y, \theta) = 1$ indicates that the attacker believes the sample $(x, y)$ was used to train the model with parameters $\theta$, and $A(x, y, \theta) = 0$ indicates when they believe it was not. 

In particular, we consider SOTA MIAs that output a confidence score, where higher values indicate stronger belief that the sample was used during training. The binary classification decision is then made by comparing this score against a threshold $\tau$:

\begin{equation}
    A_\tau(x, y, \theta) = \begin{cases}
        1 & \text{if } A(x, y, \theta) \geq \tau \\
        0 & \text{otherwise}
    \end{cases}
\end{equation}

For evaluation purposes, we consider a balanced dataset $D = D^+ \cup D^-$ where $D^+ \subseteq D_{\text{train}}$ represents member samples and $D^- \subseteq D_{\text{test}}$ represents non-member samples, with $|D^+| = |D^-|$. The attack's performance is measured using standard binary classification metrics:

\begin{itemize}
    \item True Positive Rate (TPR): The fraction of member samples correctly identified as members
    \begin{equation}
        \text{TPR}(\tau) = \frac{|\{(x,y) \in D^+ : A(x,y,\theta) \geq \tau\}|}{|D^+|}
    \end{equation}
    
    \item False Positive Rate (FPR): The fraction of non-member samples incorrectly identified as members
    \begin{equation}
        \text{FPR}(\tau) = \frac{|\{(x,y) \in D^- : A(x,y,\theta) \geq \tau\}|}{|D^-|}
    \end{equation}
\end{itemize}

Throughout this paper we run the following MIAs: 
\begin{enumerate}
    \item LOSS~\cite{yeom2018privacy}
    \item Likelihood Ratio Attack (LiRA)~\cite{carlini2022membership}
    \item Attack R~\cite{ye2022enhanced}
    \item Robust Membership Inference Attack (RMIA)~\cite{zarifzadeh2024low}
\end{enumerate}

\subsection{Attack Risk Predictors}
\label{sec:preliminaries:attack_risk_predictors}

Previous works have proposed various metrics as indicators of potential vulnerability to privacy attacks. We select several key predictors commonly cited in the literature~\cite{leemann2024is}:

\textbf{Loss} measures the discrepancy between a model's prediction and the ground truth label and is used by many MIAs~\cite{ye2022enhanced, shokri2017membership, liu2022membership}.

\textbf{Confidence} quantifies the margin between the model's highest probability prediction and its second-highest prediction, and is also used as the basis of an MIA~\cite{carlini2022membership}.

\textbf{Input gradient norm} captures the sensitivity of the model's output with respect to small perturbations in the input features and can be used as a form of feature attribution~\cite{leemann2024is}.

\textbf{Parameter gradient norm} evaluates how changes to the model's parameters influence its predictions. Gradient norms are leveraged by the GLiR Attack~\cite{leemann2023gaussianmembershipinferenceprivacy} and have been explored for links to vulnerability~\cite{leemann2024is}.

\textbf{SHAP Values}~\cite{lundberg2017unifiedapproachinterpretingmodel} leverages game theory to determine feature importance by measuring each input's marginal contribution to the model's predictions and has been considered in the context of understanding the privacy risks of model explanations~\cite{10646875}.

\section{Method}
\label{sec:method}
\begin{figure}
    \centering
    \subfloat[Loss traces of 3 samples from the ``frog'' class in CIFAR-10 representing different example hardness.]{\includegraphics[width=0.5\textwidth]{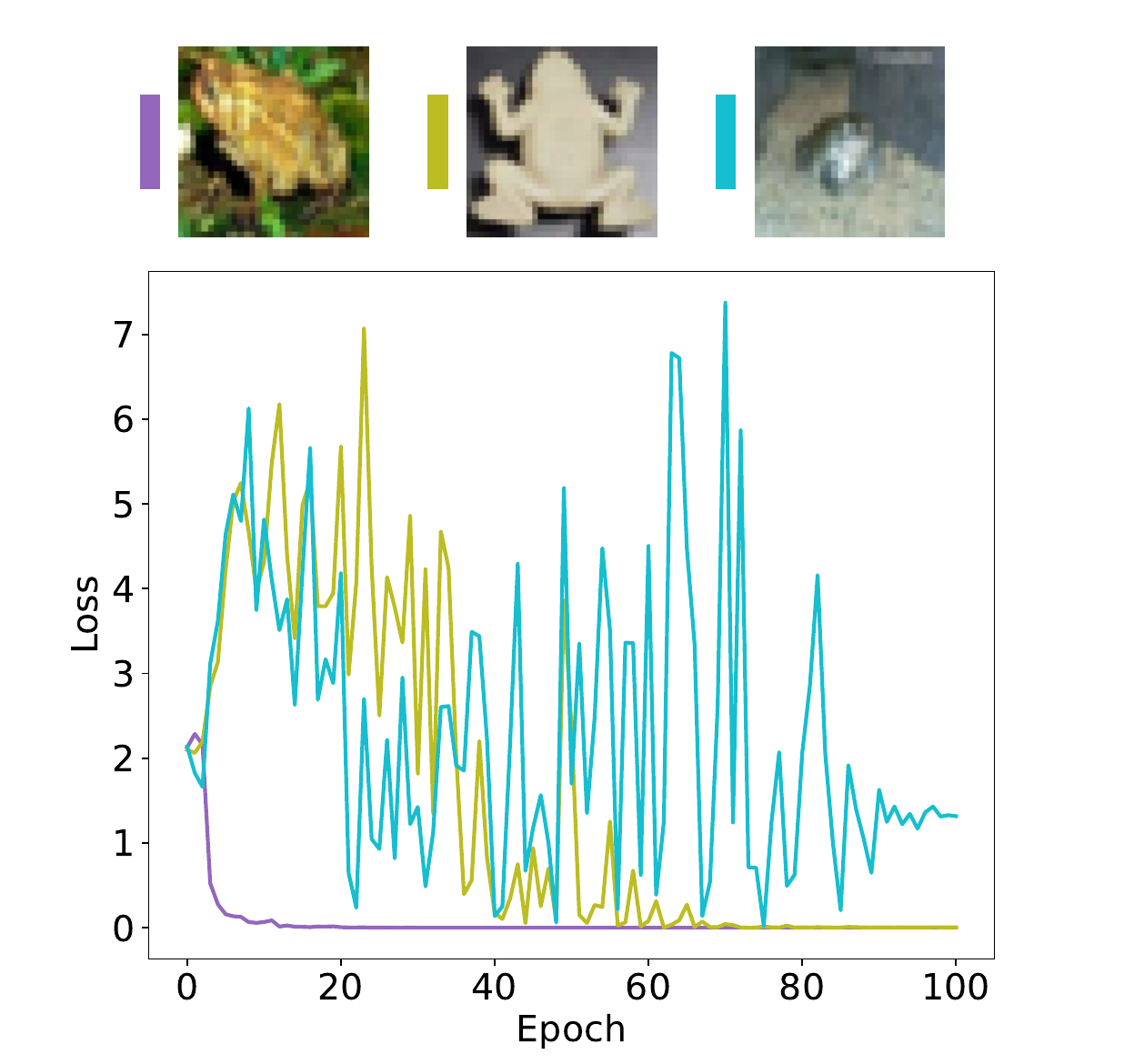}}
    \qquad
    \subfloat[Average loss traces of 2 groups of 100 points representing different example hardness and the average proportion of vulnerable points per group, compared to the average test loss.]{\includegraphics[width=0.45\textwidth]{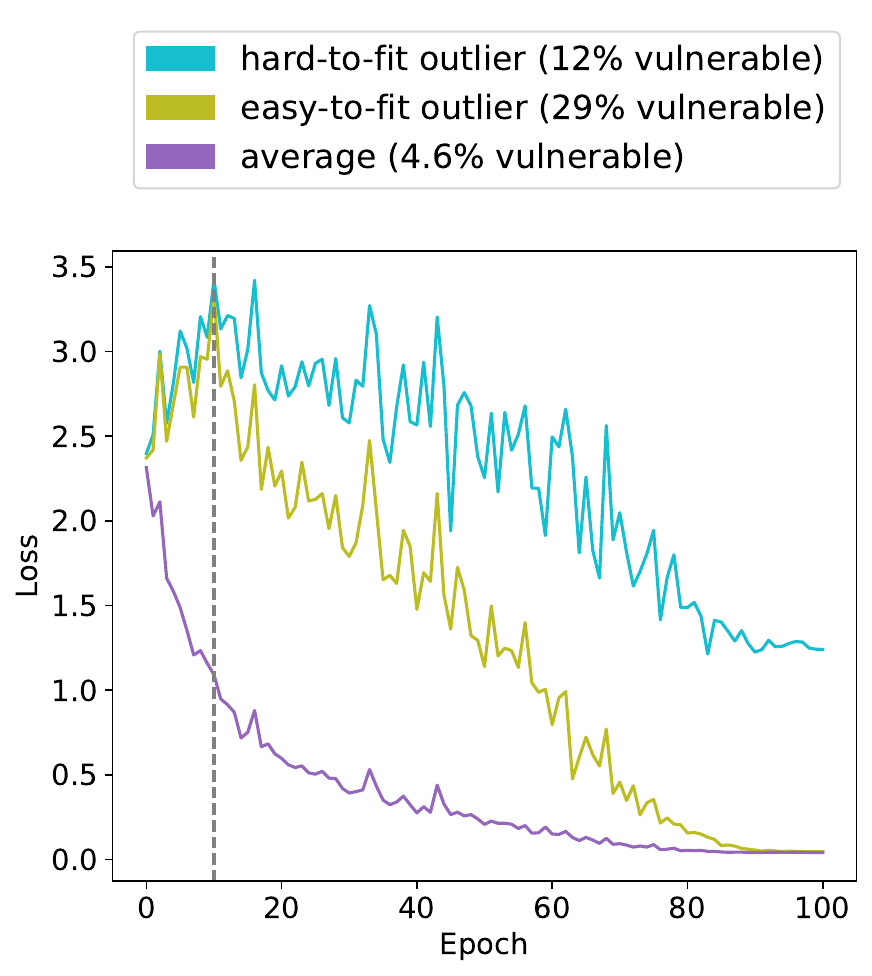}}
    \caption{Plots illustrating the intuition behind loss traces from CIFAR-10 dataset}
    \label{fig:intuition}
\end{figure}

\subsection{Intuition}
\label{sec:method:intuition}

The main challenge of running loss-based MIAs stems from the fact that the final loss for a given sample is influenced by two independent factors: \textit{example hardness} and \textit{memorization}~\cite{carlini2022membership}. For instance, extremely low loss values can either represent a very easy-to-classify example or a well-memorized one. Similarly, a relatively high loss value can represent either a hard-to-classify, partially memorized example, or an average non-memorized example.

MIAs relying on shadow models~\cite{carlini2022membership,ye2022enhanced} aim to account for example hardness by training a number of models on datasets without the target sample. The resulting loss distributions of the target sample are then measured, and used to inform the attack.

We here posit that example hardness might also be inferred, for free, from training artifacts such as the loss trace over training, and then used to evaluate the corresponding privacy risk. The intuition is that easy-to-fit samples would exhibit an early drop in the training loss and quickly reach a small loss value. Meanwhile, hard-to-classify yet memorized samples would take more epochs to converge to a final, and equally small, loss value. A similar logic applies to samples with moderately high losses at the end of training. If they are very hard-to-classify samples exhibiting even higher losses on non-member shadow models, they could still be correctly identified by an MIA. Here, a steep drop in loss during training (even if not reaching zero) would signal that a sample is partially memorized.

Fig.~\ref{fig:intuition}(a) shows loss traces on three examples from the CIFAR-10 ``frog'' class: an \textit{average} example, an \textit{easy-to-fit outlier} and a \textit{hard-to-fit outlier}. Loss traces were collected during training, and all samples were part of the training dataset. These traces represent distinct types of loss behavior.  The \textcolor{purple}{\textbf{purple}} sample (\textit{average}) is a typical sample for the frog class: it is easily learned by the model and the train loss quickly converges to small loss values. The \textcolor{olive}{\textbf{olive}} sample (\textit{easy-to-fit outlier}), on the other hand, features unique characteristics compared to other samples from the same class, notably the color. Here we can see that the loss is quite noisy and remains fairly high until approximately epoch 70 when it gets memorized.  Finally, the \textcolor{cyan}{\textbf{cyan}} sample (\textit{hard-to-fit outlier}) is an even harder-to-predict sample from the frog class. The early loss (around epoch 11) for this sample is significantly higher than for the purple example. The loss then decreases, but does not reach the low values exhibited by the purple and olive samples by the end of training. Here, the purple example would be the least vulnerable, and the olive example would be the most vulnerable.

Following this intuition, we want to further verify our hypothesis that easy-to-fit outliers would be the most vulnerable, followed by hard-to-fit outliers, and the average samples would be the least vulnerable. To confirm this, we collect 100 samples from each group and measure the proportion of samples considered vulnerable by LiRA~\cite{carlini2022membership} at $\alpha = 10^{-3}$ (see Def.~\ref{def:vulnerable}) in these groups. In particular, on Figure~\ref{fig:intuition}(b) we consider average loss traces for two groups with 100 samples each, alongside the average train loss. Both groups are selected to have relatively high loss at an early sample in the training (epoch 11, vertical dashed line on the graph) making them more likely to be outliers and at-risk samples. One group then converges to near-zero training loss, representing easy-to-fit outliers. The final loss for the final group remains relatively high, thus representing hard-to-fit outliers. Figure~\ref{fig:intuition}(b) indeed confirms that vulnerable samples make up 29\% of the easy-to-fit outlier group compared to 12\% for hard-to-fit outliers. When comparing to the average baseline of 5\%, we confirm that samples with high loss at the early stage of training, outliers, are more vulnerable to MIAs and that, among outliers, easy-to-fit samples (defined by the low final loss) are more vulnerable than hard-to-fit ones. Our method aims to utilize this information to identify at-risk samples.

\subsection{Loss trace aggregation}
\label{sec:method:trace_aggregation}

For each training sample $(x,y) \in D_\text{train}$ we consider a sequence of per-epoch losses (see Sec.~\ref{sec:preliminaries:problem_setup:model_training}):

$$
\text{trace}(x,y) = [\ell_0(x,y), \ell_1(x,y), \dots, \ell_S(x,y)]
$$ where $S$ is the total number of training epochs.

To make use of the available training artifacts and keep the additional computations to a minimum, we approximate $\ell_i(x,y)$ (defined in Sec.~\ref{sec:preliminaries:problem_setup:model_training} as the loss \textit{after} the $i$-th epoch) as the training loss computed \textit{during} the $i$-th epoch. We then consider multiple aggregating functions to compute the \textit{vulnerability score} $\phi$. 

Note that, where relevant, we use \textit{early epoch} $s^*$ to refer to an epoch at which the average test loss $\mathcal{L}_\text{test}^{s^*}$ has decreased significantly compared to the initial $\mathcal{L}_\text{test}^0$, but the train/test gap ($\mathcal{L}_\text{test}^{s^*} - \mathcal{L}_\text{train}^{s^*}$) remains low. Intuitively, we consider epoch $s^*$ as a point where the model has gained some generalization ability, but has not significantly memorized any samples.

Below we discuss various aggregation methods on the loss trace.

\paragraph{\textbf{Mean loss (LT-Mean)}}

\begin{equation}
\phi(x, y) = \frac{1}{S} \sum_{s=1}^S \ell_s(x, y) 
\end{equation}

\paragraph{\textbf{Loss Trace $L_p$ norms (LT-LP-Norm)}}

\begin{equation}
    \phi(x,y) = ||\text{trace}(x,y)||_p
\end{equation}

\paragraph{\textbf{Loss Trace Slope (LT-Slope)}} Via least squares, we compute a linear regression on the time steps against the trace, taking the gradient. This describes the rate at which the sample is learned.

\begin{equation}
    \phi(x,y) = 
    \frac{\Sigma_{s=1}^S(\ell_s(x, y)- \overline{\ell(x,y)})(s - \overline{s})}
    {\Sigma_{s=1}^S(s - \overline{s})^2}
\end{equation}

\paragraph{\textbf{Loss Trace Delta (LT-Delta)}} This represents how much the model has learned about this particular point.

\begin{equation}
\phi(x, y) = \ell_{s^*}(x, y) - \ell_S(x, y)
\end{equation}

\paragraph{\textbf{Loss Trace IQR (LT-IQR)}} This robustly measures how much a sample's loss fluctuates during training by computing the spread between two chosen quantiles of its loss trace, capturing the volatility in how well the model learns the specific example over time. Below $Q_{q_1}$ and $Q_{q_2}$ are the $q_1$-th and $q_2$-th quantiles of the loss values across all epochs, with $0 \leq q_1 < q_2 \leq 1$

\begin{equation}
\phi(x, y) = Q_{q_2}(\{\ell_s(x, y)\}_{s=1}^S) - Q_{q_1}(\{\ell_s(x, y)\}_{s=1}^S)
\end{equation}

\section{Evaluation}
\label{sec:evaluation}
\subsection{Experimental setup}
\label{sec:evaluation:setup}

\subsubsection{Target model} 
\label{sec:evaluation:setup:target_model}

Adopting the evaluation setup from the literature~\cite{carlini2022membership}, we train WideResNet28-2~\cite{BMVC2016_87} models on the CIFAR-10 dataset~\cite{Krizhevsky09learningmultiple} for 100 epochs, reaching a test accuracy of $\sim$91\%. We train using a stochastic gradient descent optimizer, with a momentum of 0.9 and weight decay set to 0.0005. We use a cosine annealing learning rate scheduler with initial value of 0.1 and a batch size of 256. During training, we augment the images via random horizontal flips, and random crops with four pixels of padding, following the augmentation setup used for training by Carlini et al.~\cite{carlini2022membership}.

We randomly split the training part of CIFAR-10 into member and non-member subsets, each containing 25000 samples. We then use the member subset for training.

We track the per-sample losses every epoch. As we use augmentations, we collect training losses on the non-augmented samples during the evaluation loop after each epoch.

For additional experiments, we vary the model or the dataset used. This involves training WideResNet28-2 models on CIFAR-100~\cite{Krizhevsky09learningmultiple} \& CINIC10~\cite{darlow2018cinic10imagenetcifar10}, as well as training ResNet-20~\cite{he2016deepresidual} \& WideResNet40-4~\cite{BMVC2016_87} models on CIFAR-10. In these cases, the training setup is otherwise as previously detailed. In all setups, the train accuracy exceeds 99\%.

\subsubsection{Membership Inference Attacks} 
\label{sec:evaluation:setup:mias}

We evaluate four distinct membership inference attacks: LiRA, Attack R, RMIA and LOSS. We follow the setup proposed in Carlini et al.~\cite{carlini2022membership}, instantiating online LiRA with 256 shadow models (128 member and 128 non-member models respectively). We instantiate online RMIA with 256 shadow models, and instantiate Attack R with 128 non-member shadow models. All shadow models are trained using the same training procedure described for the target model in Sec.~\ref{sec:evaluation:setup:target_model}, with each shadow model trained on a different member/non-member split. For LiRA, we use 2 queries per sample through horizontal flipping, matching their ``mirror augmentation'' setup.

\begin{figure}
    \centering
    \includegraphics[width=0.4\textwidth]{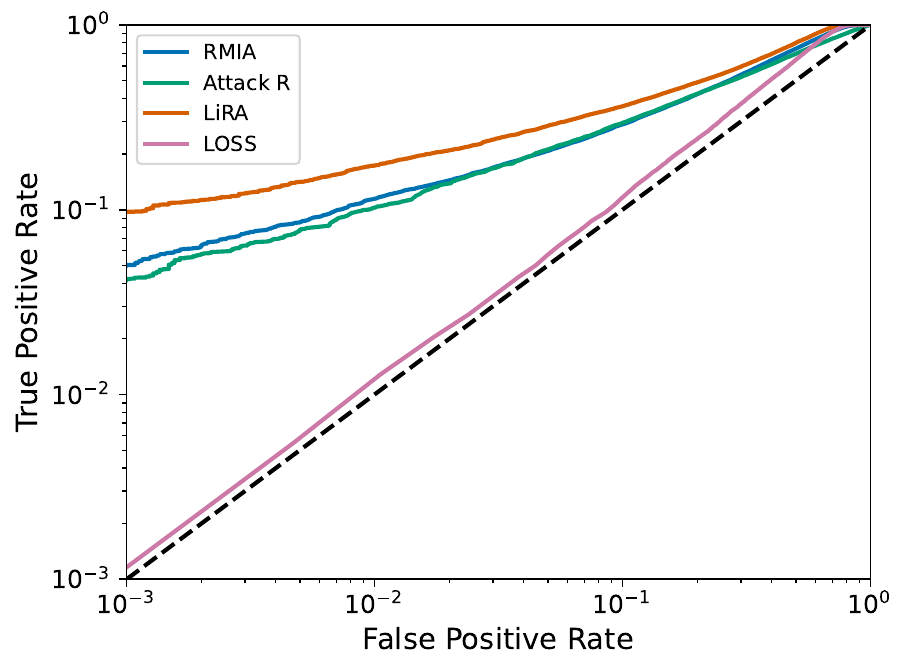}
    \caption{ROC curves for LiRA, Attack R, RMIA and LOSS membership inference attacks on CIFAR-10 against WideResNet28-2 (log scale).}
    \label{fig:roc_curves}
\end{figure}

Table~\ref{tab:all_attacks_performance} shows the performance of MIAs we consider in our training setup. We evaluate them on a random balanced split of members and non-members from the CIFAR-10 train split. As expected, SOTA shadow model-based attacks (LiRA, Attack R, RMIA) significantly outperform LOSS, with LiRA showing the best performance. Note however that though the LOSS attack shows non-trivial AUC=$0.61$, it fails to confidently identify members - as shown by the near-random chance TPR at low FPR values (see Fig.~\ref{fig:roc_curves}).

\subsection{Vulnerability to LiRA}
\label{sec:evaluation:main_results}

\begin{table*}[ht]
    \centering
    \begin{tabular}{c|cc|cc|cc}
    \toprule
        & \multicolumn{2}{c|}{k = 1\%} & \multicolumn{2}{c|}{k = 3\%} & \multicolumn{2}{c}{k = 5\%} \\
        Method & Precision@k & Recall@k & Precision@k & Recall@k & Precision@k & Recall@k \\
        \midrule
        \midrule
        Loss & 0.21 & 0.021 & 0.20 & 0.060 & 0.20 & 0.102 \\
        \midrule
        Confidence & 0.20 & 0.020 & 0.18 & 0.056 & 0.20 & 0.100 \\
        \midrule
        Parameter Gradient Norm & 0.20 & 0.020 & 0.19 & 0.059 & 0.21 & 0.108 \\
        \midrule
        Input Gradient Norm & 0.20 & 0.020 & 0.20 & 0.060 & 0.21 & 0.106 \\
        \midrule
        SHAP Values & 0.13 & 0.013 & 0.14 & 0.043 & 0.14 & 0.071 \\
        \midrule
        \textbf{LT-IQR} & \bm{\mathrm{0.92}} & \bm{\mathrm{0.094}} & \bm{\mathrm{0.83}} & \bm{\mathrm{0.26}} & \bm{\mathrm{0.76}} & \bm{\mathrm{0.39}} \\
        \bottomrule
    \end{tabular}
    \caption{Precision@k and Recall@k on identifying points vulnerable to LiRA attack. Reported at fixed FPR=$10^{-3}$ and varying values of $k$ for WideResNet28-2 on CIFAR-10}
    \label{tab:lira_precision_vary_k}
\end{table*}

We first compare our method to attack risk predictors from the literature (See Sec.~\ref{sec:preliminaries:attack_risk_predictors}). We show our method to successfully identify the points most vulnerable to the LiRA MIA and strongly outperform prior methods. Table~\ref{tab:lira_precision_vary_k} shows that at a fixed false positive rate of $10^{-3}$, LT-IQR achieves remarkable precision of 92\% at $k=1$\% (250 samples), substantially outperforming conventional approaches such as loss or gradient norm. This performance advantage is maintained even as we increase $k$ to include a larger portion of training samples, with LT-IQR maintaining 76\% precision at k=5\% while other methods remain around 20\%.

From a recall perspective, LT-IQR demonstrates consistent improvement as k increases, rising from $9\%$ at $k=1\%$ to $39\%$ at $k=5\%$ - while maintaining a relatively high precision of $76\%$. This indicates that our method successfully identifies a significant portion of truly vulnerable samples when allowed to make more predictions. In contrast, traditional metrics show minimal recall improvement with increasing $k$, never exceeding $11\%$ even at $k=5\%$.

The stark performance gap between LT-IQR and traditional metrics highlights the importance of considering training dynamics rather than just final model states when assessing sample vulnerability. Single-point metrics like final loss or gradient norms, despite their widespread use in privacy analysis, appear to capture limited information about a sample's true vulnerability to membership inference attacks. SHAP values, which attempt to quantify feature importance, perform particularly poorly with precision below $0.15$, suggesting that feature attribution alone is insufficient for identifying vulnerable samples.

\begin{table*}[ht]
    \centering
    \begin{tabular}{c|cc|cc|cc}
    \toprule
        & \multicolumn{2}{c|}{RN-20} & \multicolumn{2}{c|}{WRN28-2} & \multicolumn{2}{c}{WRN40-4} \\
        Method & Precision@k & Recall@k & Precision@k & Recall@k & Precision@k & Recall@k \\
        \midrule
        \midrule
        Loss & 0.21 & 0.039 & 0.21 & 0.021 & 0.11 & 0.012 \\
        \midrule
        Confidence & 0.19 & 0.035 & 0.20 & 0.020 & 0.10 & 0.011 \\
        \midrule
        Parameter Gradient Norm & 0.17 & 0.031 & 0.20 & 0.020 & 0.12 & 0.013 \\
        \midrule
        Input Gradient Norm & 0.19 & 0.035 & 0.20 & 0.020 & 0.12 & 0.014\\
        \midrule
        SHAP Values & 0.06 & 0.011 & 0.13 & 0.014 & 0.15 & 0.017 \\
        \midrule
        \textbf{LT-IQR} & \bm{\mathrm{0.79}} & \bm{\mathrm{0.145}} & \bm{\mathrm{0.92}} & \bm{\mathrm{0.094}} & \bm{\mathrm{0.91}} & \bm{\mathrm{0.105}} \\
        \bottomrule
    \end{tabular}
    \caption{Precision@k and Recall@k on identifying points vulnerable to LiRA attack. Reported at fixed FPR=$10^{-3}$ and $k=1\%$ for varying model architectures on CIFAR-10} 
    \label{tab:lira_precision_vary_model}
\end{table*}

\begin{table*}[ht]
    \centering
    \begin{tabular}{c|cc|cc|cc}
    \toprule
        & \multicolumn{2}{c|}{CIFAR-10} & \multicolumn{2}{c|}{CIFAR-100} & \multicolumn{2}{c}{CINIC-10} \\
        Method & Precision@k & Recall@k & Precision@k & Recall@k & Precision@k & Recall@k \\
        \midrule
        \midrule
        Loss & 0.21 & 0.021 & 0.27 & 0.011 & 0.30 & 0.023 \\
        \midrule
        Confidence & 0.20 & 0.021 & 0.24 & 0.009 & 0.28 & 0.021 \\
        \midrule
        Parameter Gradient Norm & 0.20 & 0.020 & 0.22 & 0.009 & 0.29 & 0.022 \\
        \midrule
        Input Gradient Norm & 0.20 & 0.020 & 0.23 & 0.009 & 0.30 & 0.023 \\
        \midrule
        SHAP Values & 0.13 & 0.014 & 0.18 & 0.007 & 0.23 & 0.018 \\
        \midrule
        \textbf{LT-IQR} & \bm{\mathrm{0.92}} & \bm{\mathrm{0.094}} & \bm{\mathrm{0.97}} & \bm{\mathrm{0.040}} & \bm{\mathrm{0.94}} & \bm{\mathrm{0.072}} \\
        \bottomrule
    \end{tabular}
    \caption{Precision@k and Recall@k on identifying points vulnerable to LiRA attack. Reported at fixed FPR=$10^{-3}$ and $k=1\%$ for varying datasets using WideResNet28-2}
    \label{tab:lira_precision_vary_dataset}
\end{table*}

\begin{table}[ht]
    \centering
    \begin{tabular}{c|c|cc}
    \toprule
        Dataset & Model & AUC & TPR@FPR=$10^{-3}$ \\
        \midrule
        \multirow{3}{*}[-1ex]{CIFAR-10} & RN-20 & 0.70 & 0.054 \\
        \cmidrule{2-4}
        & WRN40-4 & 0.73 & 0.086 \\
        \cmidrule{2-4}
        & \multirow{3}{*}[-1ex]{WRN28-2} & 0.74 & 0.097 \\
        \cmidrule{1-1}
        \cmidrule{3-4}
        CIFAR-100 & & 0.92 & 0.240\\
        \cmidrule{1-1}
        \cmidrule{3-4}
        CINIC-10 & & 0.80 & 0.129\\
        \bottomrule
    \end{tabular}
    \caption{LiRA performance on various models and datasets} 
    \label{tab:lira_auc_and_tpr}
\end{table}

To validate the robustness of our method, we evaluate LT-IQR across different model architectures and datasets. Note that the underlying LiRA attack shows varying base effectiveness across these configurations (shown in Table~\ref{tab:lira_auc_and_tpr}), resulting in a different number of vulnerable points for each combination of the model and the dataset.

First, we run the evaluation on CIFAR-10 dataset with three model architectures: ResNet-20 (RN-20), WideResNet28-2 (WRN28-2), and WideResNet40-4 (WRN40-4), with results presented in Table~\ref{tab:lira_precision_vary_model}. Across architectural variants LT-IQR maintains strong performance, achieving $79\%$, $92\%$, and $91\%$ precision respectively at $k=1\%$, while traditional metrics remain around $20\%$.

Second, we run the evaluation with WideResNet28-2 on three datasets: CIFAR-10, CIFAR-100, and CINIC-10 (Table~\ref{tab:lira_precision_vary_dataset}). LT-IQR maintains its discriminative power - achieving even higher precision of $97\%$ on CIFAR-100 and $94\%$ on CINIC-10. While recall appears lower on CIFAR-100 ($0.040$ vs $0.094$ on CIFAR-10), this is expected as the stronger attack identifies a larger set of vulnerable points, making it harder to capture a fixed fraction with the same $k$.

We present further results for varying values of $k$ in Appendix~\ref{app:results}.

\begin{figure}
    \centering
    \includegraphics[width=0.35\textwidth]{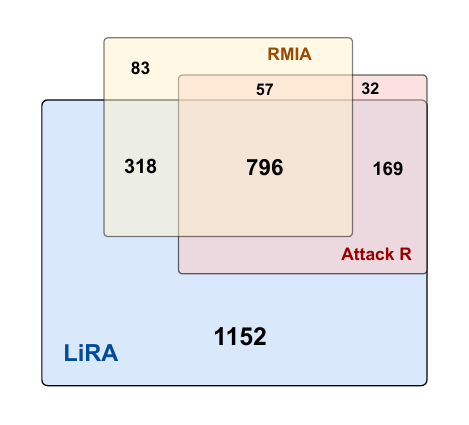}
    \caption{Venn diagram comparing the overlap between the set of points predicted to be vulnerable by LiRA, Attack R and RMIA at FPR=$10^{-3}$.}
    \label{fig:venn}
\end{figure}

\subsection{Cross-attack vulnerability}
\label{sec:evaluation:attack_union}
So far we have only considered vulnerability to one attack, LiRA. It's worth noting, however, that when evaluating privacy risks, model developers cannot predict which specific membership inference attack an adversary might use. While LiRA generally shows the strongest performance and would be a natural choice for attackers, points that are confidently identified as members by other attacks, such as Attack R or RMIA, also represent significant privacy risks.

To understand the relationships between different attacks' vulnerabilities, we first analyze how much agreement exists between state-of-the-art attacks in identifying vulnerable points. We consider three widely adopted shadow model-based MIAs: LiRA, Attack R, and RMIA, examining the overlap in points they identify as vulnerable at FPR=$10^{-3}$.

Our analysis reveals that LiRA's coverage is quite comprehensive: approximately 93\% of all points identified as vulnerable by any attack are also identified by LiRA (see Fig.~\ref{fig:venn}). However, there remains a notable 7\% of vulnerable points that are only identified by Attack R and/or RMIA. This indicates that while LiRA is highly effective, focusing solely on LiRA vulnerability could miss some genuinely at-risk samples.

Based on this observation, we expand our definition of vulnerable points to include any training sample identified as vulnerable by \textit{any} of these three attacks. We define this set as the union of vulnerable points across all attacks. As shown in Table~\ref{tab:union}, our LT-IQR method maintains its strong performance when evaluated against this broader vulnerability definition, achieving Precision@k of $0.93$ for identifying points vulnerable to any attack.

Remarkably, Table~\ref{tab:all_attacks_performance} shows that when evaluated with this expanded definition of vulnerability, our method's performance is competitive with shadow model-based approaches themselves. At $k = 1\%$, LT-IQR's $0.93$ precision is comparable to RMIA with 2 shadow models ($0.96$) and outperforms Attack R with 8 models ($0.74$), while requiring only a fraction of their computational cost.

\begin{table}[ht]
    \centering
    \begin{tabular}{c|cc}
    \toprule
        & \multicolumn{2}{c}{Union}\\
        Method & Precision@k & Recall@k \\
        \midrule
        \midrule
        Loss (desc) & 0.21 & 0.020 \\
        \midrule
        Confidence & 0.20 & 0.019 \\
        \midrule
        Parameter Gradient Norm & 0.20 & 0.020\\
        \midrule
        Input Gradient Norm & 0.20 & 0.020\\
        \midrule
        SHAP Values & 0.17& 0.016\\
        \midrule
        \textbf{LT-IQR} & \bm{\mathrm{0.93}} & \bm{\mathrm{0.089}}  \\
        \bottomrule
    \end{tabular}
    \caption{Precision@k and Recall@k at predicting the vulnerable points defined by the union of three attacks: LiRA, Attack R and RMIA. Reported at fixed FPR=$10^{-3}$ and varying values of $k$ for WideResNet28-2 on CIFAR-10}
    \label{tab:union}
\end{table}

\begin{table*}[ht]
    \centering
    \begin{tabular}{c|c|cc|p{1.5cm}p{1.5cm}p{1.5cm}l}
    \toprule
        & & \multicolumn{2}{c|}{MIA performance} & \multicolumn{3}{c}{Precision@k} & \\
        Method & \shortstack{\# Shadow\\models} & AUC & \shortstack{TPR@FPR=$10^{-3}$} & k=1\% & k=3\% & k=5\%  &k=10\%\\
        \midrule
        \midrule
        \multirow{3}{*}{LiRA} & 256 & 0.74 & 0.097& 1.00 & 1.00 & 1.00 &0.98\\
        \cmidrule{2-8}
        & 64 & 0.73 & 0.066& 1.00 & 1.00 & 0.99 &0.89\\
        \cmidrule{2-8}
        & 16 & 0.69 & 0.046& 0.99 & 0.97 & 0.92 &0.76\\
        \midrule
        \multirow{3}{*}{Attack R} & 128 & 0.66 & 0.042& 1.00 & 1.00 & 0.97 &0.78\\
        \cmidrule{2-8}
        & 32 & 0.66 & 0.000& 0.91 & 0.97 & 0.94 &0.76\\
        \cmidrule{2-8}
        & 8 & 0.65 & 0.000& 0.74 & 0.90 & 0.90 &0.75\\
        \midrule
        \multirow{4}{*}{RMIA} & 256 & 0.69& 0.050& 1.00 & 1.00& 1.00&0.77\\
        \cmidrule{2-8}
        & 64 & 0.69& 0.052& 1.00 & 1.00& 0.97&0.75\\
        \cmidrule{2-8}
        & 16 & 0.67 & 0.045& 1.00 & 1.00& 0.95&0.74\\
        \cmidrule{2-8}
        & 2 & 0.65 & 0.026& 0.96& 0.87& 0.80&0.62\\
        \midrule
        LOSS & -- & 0.61 & 0.000& 0.11& 0.07 & 0.05&0.05\\
        \midrule
        \midrule
        LT-IQR & -- & -- & --& 0.93 & 0.83 & 0.77 &0.62\\
        \bottomrule
    \end{tabular}
    \caption{Precision@k for traditional MIAs identifying the union of points vulnerable to at least one of the shadow model-based MIAs (LiRA, Attack R, RMIA) with varying number of shadow models. Reported at fixed FPR=$10^{-3}$ on CIFAR-10 with WideResNet28-2)}
    \label{tab:all_attacks_performance}
\end{table*}

\subsection{Loss trace aggregation methods}
\label{sec:evaluation:loss_aggregations}

\begin{table*}[ht]
    \centering
    \begin{tabular}{c|cc|cc|cc}
    \toprule
        & \multicolumn{2}{c|}{CIFAR-10} & \multicolumn{2}{c|}{CIFAR-100} & \multicolumn{2}{c}{CINIC-10} \\
        Method & Precision@k & Recall@k & Precision@k & Recall@k & Precision@k & Recall@k \\
        \midrule
        \midrule
        LT-Mean & $0.91$ & $0.093$ & $0.96$ & $0.040$& \bm{\mathrm{0.96}} & \bm{\mathrm{0.074}}\\
         \midrule
        LT-Delta & $0.51$ & $0.052$ & $0.62$& $0.026$& $0.78$&$0.060$ \\
         \midrule
        LT-Slope & $0.89$ & $0.091$ & \bm{\mathrm{0.98}}& \bm{\mathrm{0.040}} & $0.92$&$0.070$ \\
        \midrule
        LT-L2-Norm & $0.87$ & $0.089$ & $0.97$& $0.040$& $0.95$&$0.073$ \\
         \midrule
        LT-Linf-Norm & $0.42$ & $0.042$ & $0.13$& $0.005$& $0.19$&$0.014$ \\
         \midrule
        LT-IQR & \bm{\mathrm{0.92}} & \bm{\mathrm{0.094}} & $0.97$ & $0.040$ & $0.94$ & $0.072$ \\
         \bottomrule
    \end{tabular}
    \caption{Precision@k and Recall@k for various loss trace aggregation methods, reported at fixed FPR=$10^{-3}$ and $k=1\%$ with WideResNet28-2.} 
    \label{tab:all_metrics_precision_recall}
\end{table*}

We propose a novel approach for identifying vulnerable points through the analysis of training dynamics. While we use Loss Trace Interquartile Range (LT-IQR) as our primary method in experimental evaluations, we explored several other aggregation techniques that have shown strong performance across different datasets and architectures.

We evaluate multiple approaches to aggregate the loss traces collected during training (see Sec.~\ref{sec:method:trace_aggregation}). Table~\ref{tab:all_metrics_precision_recall} presents a comparison of different aggregation methods across CIFAR-10, CIFAR-100, and CINIC-10 datasets. LT-Mean, LT-Slope, and LT-L2-Norm consistently achieve high precision above $0.90$. While LT-IQR maintains strong performance across datasets, several other methods perform comparably well.

On the other hand, the $L_\text{inf}$ norm consistently underperforms, suggesting that aggregation of the entire loss trajectory provides a more robust signal than a single point observation.

These results indicate that while LT-IQR provides a reliable approach to vulnerability detection, the temporal information in loss traces is robust to multiple aggregation methods. The success of multiple methods suggests that the specific choice of aggregation method may be less critical than the fundamental insight of leveraging training dynamics for vulnerability detection.

\subsection{Computational overhead}
\label{sec:evaluation:overhead}

Our method relies on artifacts available during training, and does not require any additional model training, making it computationally cheap to apply in practice. We here empirically confirm this by measuring the computational overhead incurred using our method and showing that collecting losses during training has a negligible impact on training time.

Table~\ref{tab:overhead} presents the average time per epoch across 10 epochs when training a WideResNet28-2~\cite{BMVC2016_87} model on the CIFAR-10 dataset on one A100 GPU with a batch size of 64. Our results show that collecting per-sample losses during training adds no statistically significant overhead. The baseline training time of $20.86 \pm 1.02$ seconds per epoch and the time with loss collection ($21.08 \pm 1.16$ seconds) have overlapping confidence intervals, confirming that tracking individual sample losses can be seamlessly integrated into the training pipeline with effectively zero cost. Even in cases where data augmentation requires an extra forward pass to compute losses on non-augmented samples, the overhead remains low. This worst-case scenario requires $24.84 \pm 1.50$ seconds per epoch, representing roughly a 20\% increase over the baseline training time. 

\begin{table}[ht]
    \centering
    \begin{tabular}{p{4cm}c}
    \toprule
        & Time per epoch (s) \\
        \midrule
        \midrule
        Normal training & $20.86 \pm 1.02$ \\
        \midrule
        Normal training \\+ collecting ``free'' losses & $21.08 \pm 1.16$ \\
        \midrule
        Normal training \\+ extra forward pass & $24.84 \pm 1.50$ \\
        \bottomrule
    \end{tabular}
    \caption{Average time per epoch (s) across 10 epochs to train a WideResNet28-2 on CIFAR-10 on an A100 GPU.}
    \label{tab:overhead}
\end{table}

\subsection{Ablation on Quantile Parameters}
\label{sec:evaluation:ablations}

The LT-IQR score relies on two quantile parameters $q_1$ and $q_2$ that determine which portions of the loss trace are used to compute the interquartile range. To thoroughly evaluate their impact, we conduct a grid search over:

\begin{itemize}
    \item $q_1$ values in [0.0, 0.50] at 0.05 intervals
    \item $q_2$ values in [0.50, 1.00] at 0.05 intervals
\end{itemize}

We fix FPR=$10^{-3}$ and repeat for k=1\%, 3\%, \& 5\% of training data for these experiments.

\begin{figure*}[t]
    \centering
    \includegraphics[width=0.95\textwidth]{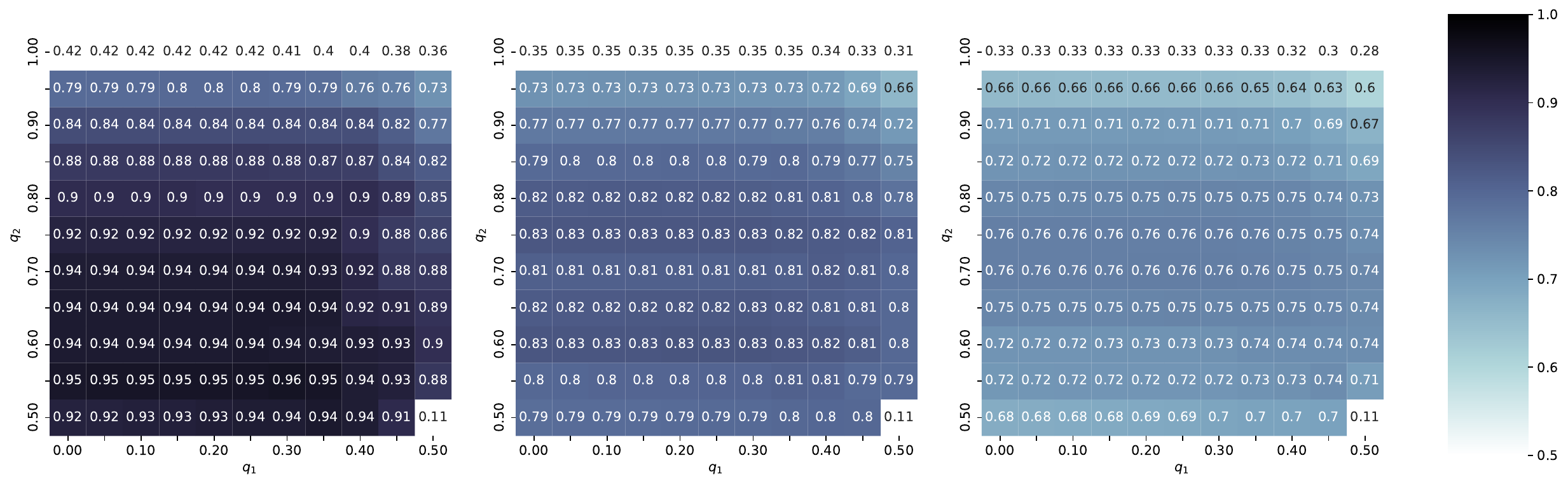}
    \caption{Heatmaps showing the Precision@k for the LT-IQR method with varying quantile parameters of $q_1$ and $q_2$. Reported at fixed FPR=$10^{-3}$ on CIFAR-10 and WideResNet28-2 for k = $1\%$ / $3\%$ / $5\%$.}
    \label{fig:iqr_heatmap}
\end{figure*}

Figure~\ref{fig:iqr_heatmap} shows a heatmap visualization of how the LT-IQR method's Precision@k varies with different quantile parameter combinations ($q_1$ and $q_2$) when evaluated on CIFAR-10 using WideResNet28-2. Within the range explored, we generally observe the value of $q_1$ to have limited impact. For $q_2$, the best performance is observed around the middle of the range or towards an earlier stage.

The heatmaps reveals that the method exhibits robust performance across a broad region of the parameter space, maintaining high precision for most reasonable parameter combinations. Based on these findings, we selected $q_1=0.25$ and $q_2=0.75$ as our default parameters, as these values lie within the high-performance region and correspond to conventional quartile definitions.

\section{Discussion}
\label{sec:discussion}
\subsection{Limitations}

While our method successfully identifies individual training samples vulnerable to membership inference attacks, some limitations and open questions remain.

\paragraph{\textbf{Calibration Challenges}}

One notable limitation of our method is the difficulty in selecting an appropriate threshold for the vulnerability score in making a binary assessment. Threshold calibration is often side-stepped in the MIA literature~\cite{carlini2022membership,sablayrolles2019white,yeom2018privacy}, with most attacks evaluated with threshold-independent methods. This lack of a principled way to calibrate scores means that practitioners might have to rely on relative rankings rather than absolute measures of vulnerability.

\subsection{Future work}

\paragraph{\textbf{Model-Level Vulnerability}}

\begin{figure}
    \centering
    \includegraphics[width=0.45\textwidth]{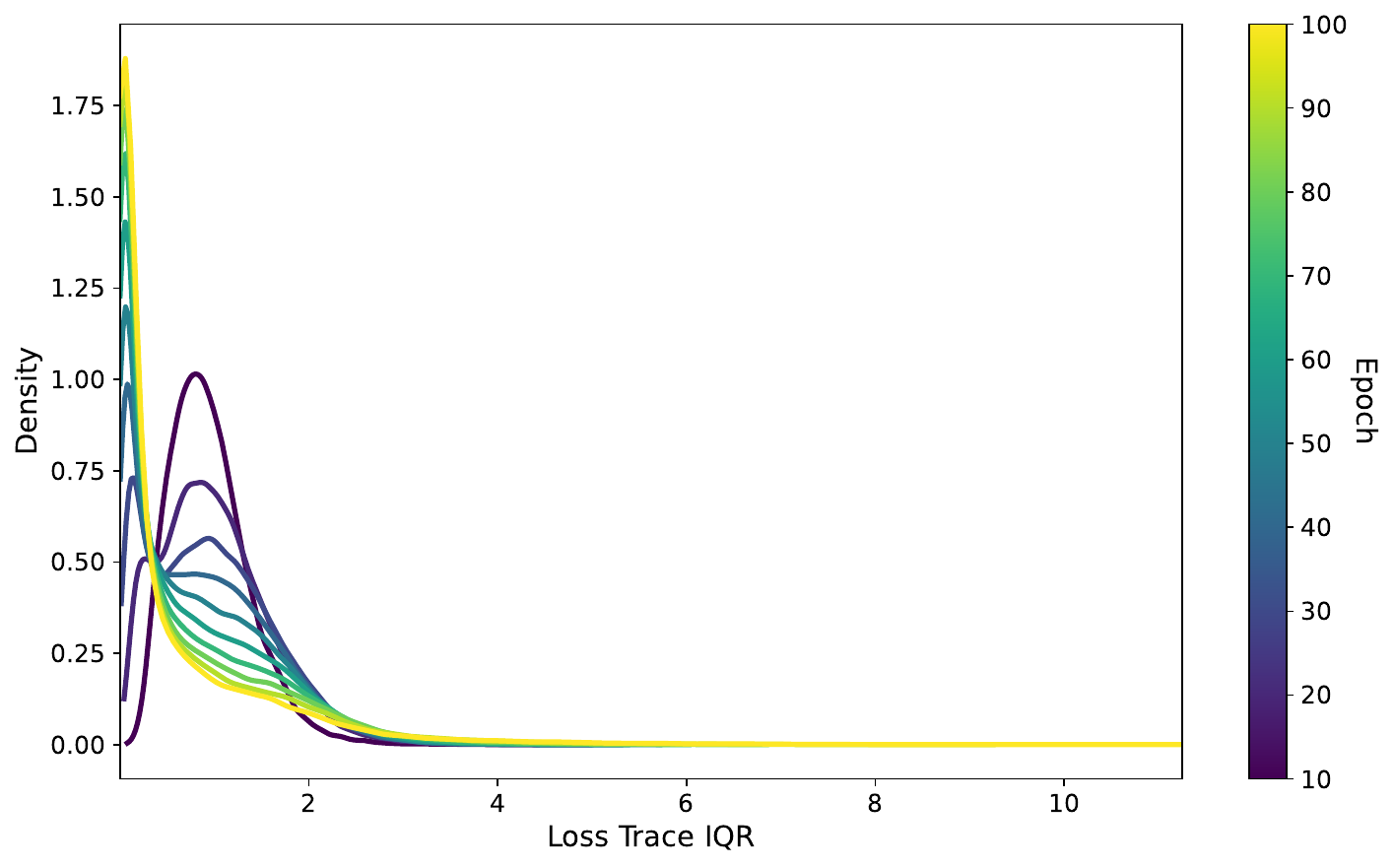}
    \caption{Distribution of sample LT-IQR scores as computed on the partial loss traces at various epochs of training. Each line represents the LT-IQR distribution on the entire training dataset after a given particular epoch.}
    \label{fig:density_plot}
\end{figure}

\begin{table}[ht]
    \centering
    \begin{tabular}{c|c|cc}
    \toprule
        Dataset & Model & \shortstack{LiRA TPR\\(FPR=$10^{-3}$)} &   \shortstack{Mean\\LT-IQR}\\
        \midrule
        \midrule
        \multirow{3}{*}[-1ex]{CIFAR-10} & RN-20 & 0.054 & 0.64\\
        \cmidrule{2-4}
        & WRN40-4 & 0.086 & 0.37 \\
        \cmidrule{2-4}
        & \multirow{3}{*}[-1ex]{WRN28-2} & 0.097 & 0.46 \\
        \cmidrule{1-1}
        \cmidrule{3-4}
        CIFAR-100 & & 0.239 & 1.59\\
        \cmidrule{1-1}
        \cmidrule{3-4}
        CINIC-10 & & 0.129 & 0.86\\
        \bottomrule
    \end{tabular}
    \caption{LiRA attack performance and mean LT-IQR values across different datasets and model architectures.} 
    \label{tab:model_level}
\end{table}

Analyzing the distribution of vulnerability scores across training epochs might provide insights into the overall model risk. Figure~\ref{fig:density_plot} visualizes the density distribution of LT-IQR scores across all training samples at different epochs.

Early on in the training (epochs 10-30), the LT-IQR distribution is relatively concentrated, with most samples showing similar levels of loss variation. This suggests that the model is still learning general patterns and has not yet begun significant memorization of individual samples. The tight distribution indicates that the model treats most samples in a similar way during this phase.

As training progresses (epochs 30-80), we observe that the distribution shifts towards the left, with more points having low LT-IQR. This makes the remaining points with high LT-IQR scores more distinguishable and vulnerable. This transformation in the distribution shape corresponds to the model beginning to memorize individual samples, particularly outliers. Intuitively, the small proportion of points that are still experiencing higher loss volatility mid-training are the ones likely to become vulnerable to membership inference attacks.

By the final stages of training (epochs 80-100), the distribution takes on a power law-like shape, with a large concentration of samples having low LT-IQR values and a sharp but smooth decay toward higher values. 

The evolution of this distribution might provide a useful signal about whether a model is vulnerable to membership inference attacks, and when it is most vulnerable. The transformation from a concentrated distribution to a power law-like shape suggests increased memorization and thus greater privacy risk.

Additionally, Table~\ref{tab:model_level} demonstrates a positive correlation between LiRA attack effectiveness and mean LT-IQR scores across model and dataset configurations. Higher attack success rates correspond with elevated mean LT-IQR values, as shown by CINIC-10 with the strongest attack performance (TPR=0.239) and highest vulnerability score (1.59). Conversely, lower attack success rates exhibit reduced mean LT-IQR values. This correlation suggests aggregate LT-IQR statistics could serve as a proxy for overall model-level privacy risk.

\paragraph{\textbf{Test Set Analysis}}

So far in this paper we have only used the individual loss traces to predict vulnerability. However, the model trainer also has access to dataset-level information such as the loss traces of other training samples, as well as loss traces for the test samples (assuming evaluation is run regularly throughout training). Intuitively, loss traces for the test samples could act as an approximation for the denominator in the LiRA attack score. Where many SOTA MIAs estimate example hardness by relying on the loss distribution from the shadow models trained on datasets excluding the target sample, we expect that one could use the distribution of the target model losses on \textit{similar} non-member samples.

\paragraph{\textbf{Threshold Calibration Methods}}

Future research might want to explore ways to better calibrate vulnerability scores, potentially by creating hybrid approaches that combine lightweight artifact analysis with selective shadow model training, or exploring the relationship between loss traces and concrete privacy risks in real-world applications. The test set can also potentially be leveraged for threshold calibration.

\paragraph{\textbf{Informed Defense Strategies }}

Our vulnerability detection approach creates new opportunities for developing more strategic privacy defenses, especially through iterative record removal techniques. Although Carlini et al.~\cite{carlini2022privacy} demonstrated that removing the most vulnerable training samples can increase vulnerability among remaining samples, the model's overall privacy risk still decreases. This finding suggests that LT-IQR-guided record removal could offer an effective and computationally practical ad-hoc defense mechanism. Through iterative identification and elimination of records with the highest vulnerability scores, model developers could reduce privacy exposure systematically. This methodology would be especially beneficial for large-scale models where conventional privacy risk assessment is computationally prohibitive, allowing practitioners to implement mitigations early in development.

\section{Conclusion}
\label{sec:conclusion}
This work introduces a novel, computationally efficient approach that leverages readily available training artifacts (specifically, per-sample loss traces collected during the training process) to identify training samples most vulnerable to membership inference attacks. We call this approach artifact-based.

We focus specifically on identifying the small subset of samples with the highest privacy risk (a task that, while more challenging, is crucial for practical privacy protection). At a strict false positive rate of 0.001, these highly vulnerable samples constitute less than 10\% of the training data but pose the most serious privacy risks, as they can be confidently identified as training set members even under conservative attack settings.

Our method analyzes loss traces collected during model training to identify these samples without requiring prohibitively expensive shadow model training. Extensive experiments on standard benchmarks demonstrate the effectiveness of our Loss Trace Interquartile Range (LT-IQR) method, significantly outperforming existing methods for an individual privacy risk prediction.

The key findings of our work include:

\begin{itemize}
    \item The proposed Loss Trace Interquartile Range (LT-IQR) method effectively identifies training samples vulnerable to membership inference attacks by analyzing per-sample loss trajectories during training, demonstrating notably higher performance compared to existing methods. For instance, it achieves remarkable Precision@k=1\%, successfully identifying samples most vulnerable to the LiRA attack.
    \item The method maintains strong performance across different model architectures (ResNet-20, WideResNet28-2, WideResNet40-4) and datasets (CIFAR-10, CIFAR-100, CINIC-10), demonstrating its robustness and generalizability.
    \item LT-IQR performs competitively with shadow model-based approaches when we use the same precision metric to measure the agreement between the attacks. It is particularly notable as LT-IQR is computationally much cheaper, as it requires no additional model training.
    \item We propose an array of methods analyzing per-sample loss traces (LT-Mean, LT-Slope, and LT-L2-Norm), all consistently achieving high precision across datasets, demonstrating that the temporal information in loss traces can be effectively captured through multiple aggregation strategies.
\end{itemize}

These results demonstrate the promising potential of artifact-based methods in privacy-aware machine learning development. The ability to identify vulnerable samples during training, without significant computational overhead and as part of standard toolboxes, enables model developers to assess privacy risks early on during model development and implement targeted mitigations such as selective unlearning or differential privacy. Morever, the results also demonstrate the utility of artifact-based methods in understanding model memorisation with potential applications such as measuring overall model risk. This is particularly valuable for large-scale models where training hundreds of shadow models would be prohibitively expensive.

\section*{Ethics considerations}

Our research aims to enable model developers to evaluate the privacy risk of trained model at low cost, more specifically here to identify training sample that are likely to be vulnerable to membership inference attacks. This enables developers to efficiently evaluate the risk before sharing or deploying a model and, if necessary, to implement defenses. Unlike membership inference attacks, our results do not directly improves the capabilities of a general adversary. While an adversary could theoretically attempt a similar approach, their effectiveness would be severely limited without access to the training artifacts that are available to model developers. The benefit of early privacy leakage detection and mitigation therefore strongly outweigh any potential risks from adversarial exploitation. Even in the unlikely scenario where a sophisticated adversary gains access to training artifacts, our approach ensures that privacy vulnerabilities can be identified and mitigated during development, substantially reducing the attack surface before models reach production.

\section*{Open science}

The code to reproduce our results is available at \url{https://github.com/computationalprivacy/loss_traces}. It includes scripts for training target and shadow models with loss tracking, along with implementations of our proposed loss aggregation methods, as well as standard MIAs. All experiments in this work were conducted on publicly accessible datasets and models.

\bibliography{bibliography}
\bibliographystyle{plain}

\newpage
\appendix

\section{Further results}
\label{app:results}
\begin{table*}[ht]
    \centering
    \begin{tabular}{c|c|cccccc|cccccc}
    \toprule
        & & \multicolumn{6}{c|}{Precision} & \multicolumn{6}{c}{Recall} \\
        & & \multicolumn{6}{c|}{} & \multicolumn{6}{c}{} \\
        & & \multicolumn{6}{c|}{$k$} & \multicolumn{6}{c}{$k$} \\
        Dataset & Model & $1\%$ & $3\%$ & $5\%$ & $10\%$ & $20\%$ & $50\%$ & $1\%$ & $3\%$ & $5\%$ & $10\%$ & $20\%$ & $50\%$ \\
        \midrule
        \multirow{3}{*}[-1ex]{CIFAR-10} & RN-20 & .79 & .64 & .55 & .39 & .25 & .11 & .15 & .36 & .51 & .71 & .92 & 1.0  \\
        \cmidrule{2-14}
        & WRN40-4 & .91 & .83 & .75 & .59 & .39 & .17 & .11 & .29 & .43 & .69 & .91 & 1.0 \\
        \cmidrule{2-14}
        & \multirow{3}{*}[-1ex]{WRN28-2} & .92 & .83 & .76 & .60 & .42 & .19 & .09 & .26 & .39 & .61 & .86 & 1.0 \\
        \cmidrule{1-1}
        \cmidrule{3-14}
        CIFAR-100 & & .97 & .94 & .90 & .83 & .71 & .46 & .04 & .12 & .19 & .34 & .59 & .95 \\
        \cmidrule{1-1}
        \cmidrule{3-14}
        CINIC-10 & & .94 & .88 & .82 & .71 & .53 & .26 & .07 & .20 & .32 & .55 & .83 & 1.0 \\
        \bottomrule
    \end{tabular}
    \caption{Precision@k and Recall@k on identifying points vulnerable to LiRA attack using LT-IQR method. Reported at fixed FPR=$10^{-3}$ and varying values of $k$ for WideResNet28-2 on CIFAR-10} 
    \label{tab:lira_precision_ltiqr_only}
\end{table*}

Table~\ref{tab:lira_precision_ltiqr_only} demonstrates LT-IQR's robust performance across varying coverage levels. At $k=10\%$ (2500 samples), the method maintains strong results across all configurations, with recall ranging from 0.34-0.71 and precision from 0.39-0.83. For instance, on CIFAR-10 with ResNet-20, we identify over 70\% of vulnerable points with 39\% precision.

At higher coverage levels ($k=20\%$), LT-IQR achieves approximately $90\%$ recall while maintaining good precision. The method reaches perfect recall (1.0) at $k=50\%$ across nearly all configurations.

\begin{figure}
    \centering
    \includegraphics[width=0.45\textwidth]{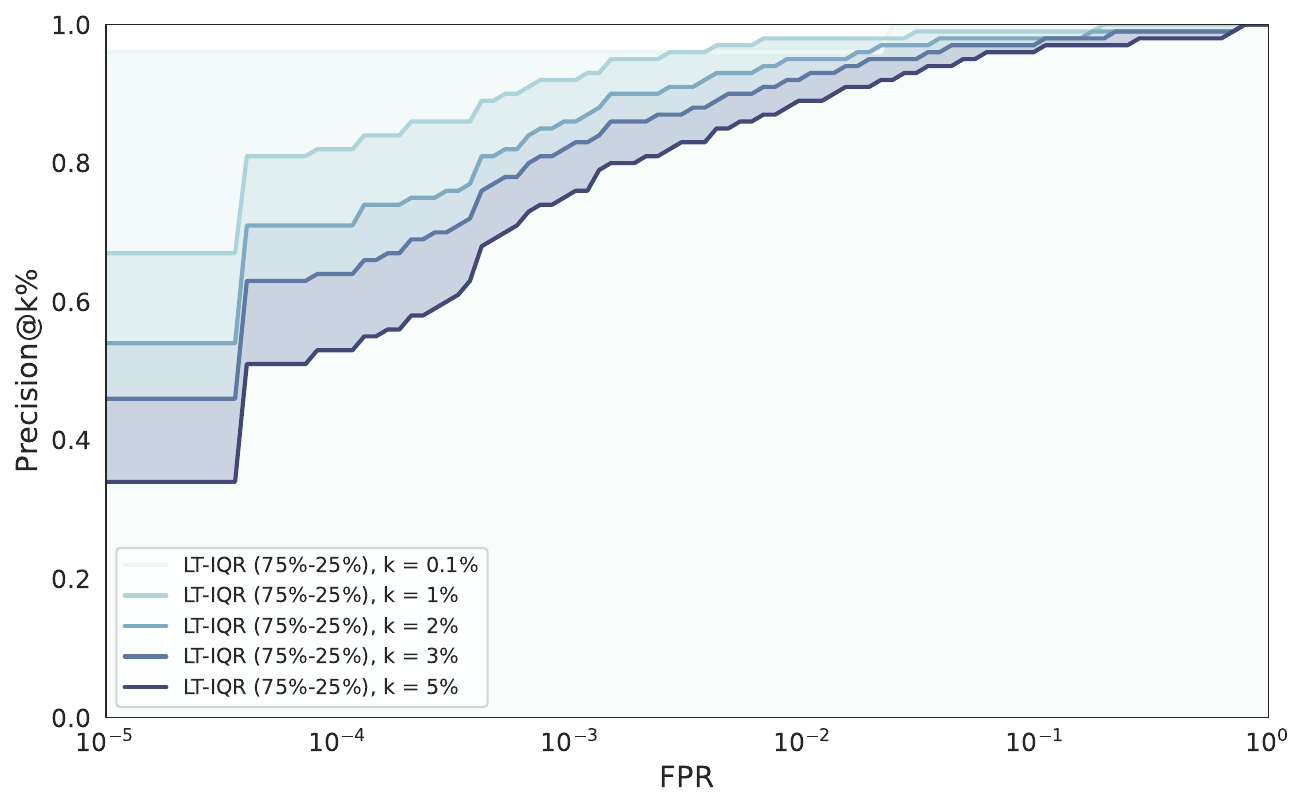}
    \caption{Precision@k of the LT-IQR method for varying values of k. Reported on CIFAR-10 and WideResNet28-2.}
    \label{fig:precision_vary_k}
\end{figure}

\begin{figure}
    \centering
    \includegraphics[width=0.45\textwidth]{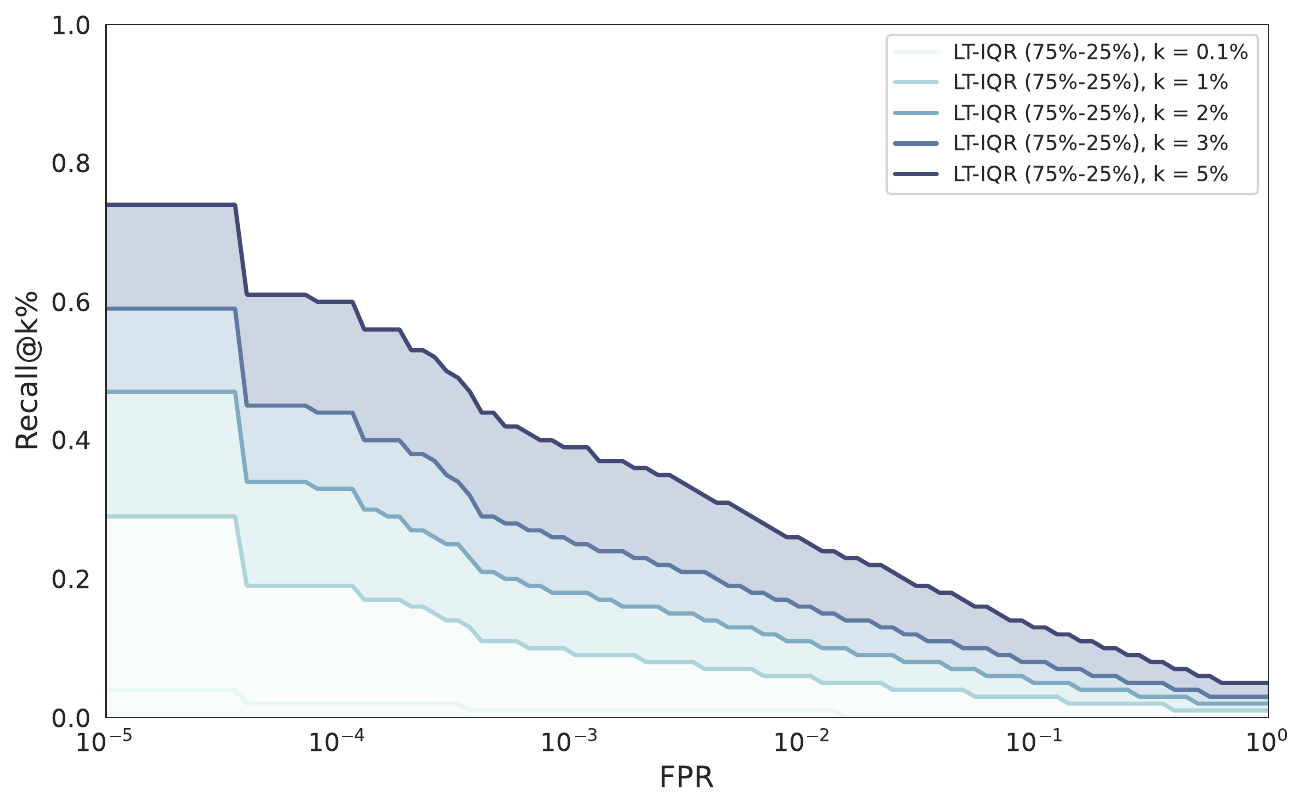}
    \caption{Recall@k of the LT-IQR method for varying values of k. Reported on CIFAR-10 and WideResNet28-2.}
    \label{fig:recall_vary_k}
\end{figure}

We then evaluate a range of $k$ values (0.1\%-5\%, corresponding to 25-1250 points respectively) across various FPR thresholds. Figure~\ref{fig:precision_vary_k} and Figure~\ref{fig:recall_vary_k} present Precision@k and Recall@k respectively.

\begin{table*}
    \centering
    \begin{tabular}{c|c|cc|cccccc}
    \toprule
        & & & &  \multicolumn{4}{c}{Max Recall@k} \\
        Dataset & Model & \shortstack{\# Member\\points} & \shortstack{\# Vulnerable\\points} & k=1\% & k=3\% & k=5\% &k=10\% &k=20\% &k=50\%\\
        \midrule
        \midrule
        \multirow{3}{*}[-1ex]{CIFAR-10} & RN-20 & 25000 & 1342 & 0.19 & 0.56 & 0.93 & 1.00 & 1.00 & 1.00 \\
        \cmidrule{2-10}
        & WRN40-4 & 25000 & 2142 & 0.12 & 0.35 & 0.58 & 1.00 & 1.00 & 1.00\\
        \cmidrule{2-10}
        & \multirow{3}{*}[-1ex]{WRN28-2} & 25000 & 2435 & 0.10 & 0.31 & 0.51 & 1.00 & 1.00 & 1.00 \\
        \cmidrule{1-1}
        \cmidrule{3-10}
        CIFAR-100 & & 25000 & 5971 & 0.04 & 0.13 & 0.21 & 0.42 & 0.84 & 1.00\\
        \cmidrule{1-1}
        \cmidrule{3-10}
        CINIC-10 & & 45000 & 5781 & 0.08 & 0.23 & 0.39 & 0.78 & 1.00 & 1.00\\
        \bottomrule
    \end{tabular}
    \caption{Maximum achievable Recall@k at different values of $k$, showing the theoretical upper bound based on the total number of vulnerable points identified by LiRA at FPR=10$^{-3}$} 
    \label{tab:max_recall}
\end{table*}

\section{Maximum achievable recall}
\label{app:recall}
Table~\ref{tab:max_recall} contextualizes our recall results by showing theoretical upper bounds determined by the ratio of $k$ samples to total number of vulnerable points. For example, on CIFAR-10 with WideResNet28-2 (2435 vulnerable points), maximum achievable recall is limited to 10\% at $k=1\%$, 31\% at $k=3\%$, with perfect recall becoming possible at k=10\%.

\section{Stability of Vulnerability Definition}
\label{app:stability}
So far we have considered a point to be vulnerable if it is confidently and correctly identified by a given MIA against the target model at a certain FPR threshold (Definition~\ref{def:vulnerable}). We now investigate whether vulnerability predictions remain stable across different training instantiations by varying the random seed used to train the target model.

\begin{definition}[Agreement-based vulnerable points]
\label{def:agreement_vulnerable}
Given $N$ instantiations of the vulnerability assessment process, each producing a set of vulnerable points $V_\alpha^{(i)}$ at FPR threshold $\alpha$, we define the set of vulnerable points requiring agreement from at least fraction $p \in [0,1]$ of instantiations as:
\begin{equation}
    V_\alpha^{(p)} = \left\{(x, y) \in D_{\text{train}} : |\{i : (x,y) \in V_\alpha^{(i)}\}| \geq \lceil p \cdot N \rceil \right\}
\end{equation}
\end{definition}

\begin{figure}
    \centering
    \includegraphics[width=0.4\textwidth]{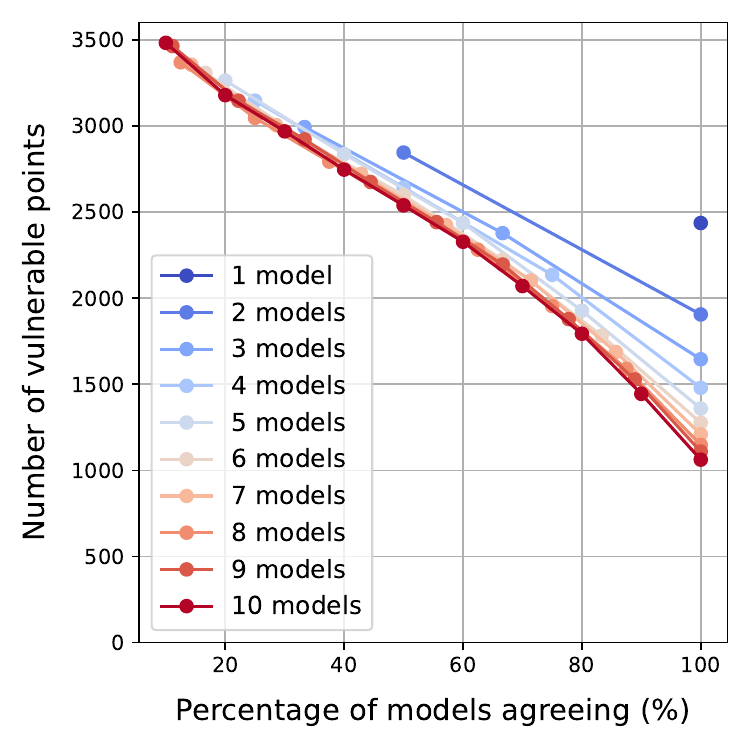}
    \caption{Number of vulnerable points according to agreement-based definition (Def.~\ref{def:agreement_vulnerable}). Lines represent varying pool sizes. Reported for WideResNet28-2, CIFAR-10, FPR=$10^{-3}$ setup.}
    \label{fig:model_agreement}
\end{figure}

We train up to 10 target models using different random seeds, independently computing vulnerable point sets for each using the full LiRA attack setup. We then consider a point vulnerable to a given MIA against a particular training process, if it is consistently identified as vulnerable for multiple instantiations of that training process.

Figure~\ref{fig:model_agreement} illustrates the relationship between vulnerable point counts, agreement percentages, and model pool sizes. Each colored line represents a different target model pool size,
ranging from 1 model (dark blue) to 10 models (red).

At the left side of each curve (low agreement percentage), we approach the union of vulnerable points across all models in the pool. For instance, with a 10\% agreement threshold and 10 models, a point is considered vulnerable if it appears in the vulnerable set of just 1 out of 10 models. This captures the broadest possible definition of vulnerability, including points that may be vulnerable due to specific quirks of individual model instantiations. 

At the right side of each curve (high agreement percentage), we approach the intersection of vulnerable points across all models. With a 100\% agreement threshold, a point is only considered vulnerable if it appears in the vulnerable set of every single model in the pool. This represents the most conservative definition, identifying only points that are consistently vulnerable regardless of training randomness. 

The single-model case (dark blue point at the top right) corresponds to our standard Definition~\ref{def:vulnerable}, where 100\% agreement from one model simply means that model identifies the point as vulnerable. As we increase the pool size, the curves become smoother and provide more granular control over the vulnerability definition. 

Importantly, the graph demonstrates convergence in the number of vulnerable points as we increase the model pool size. The curves for larger model pools (8, 9, 10 models) lie very close to each other, particularly in the middle range of agreement percentages. This convergence indicates that the vulnerability definition stabilizes and becomes robust to the specific choice of model pool size.

Table~\ref{tab:target_model_overlap} presents results using agreement-based definitions. LT-IQR maintains strong performance across all agreement thresholds, achieving 77\% precision with unanimous agreement (100\% threshold), substantially outperforming baseline methods. Interestingly, LT-IQR performs better at identifying agreement-based vulnerable points: at 50\% agreement it achieves 96\% precision compared to 92\% with standard definition at similar recall values.

\begin{table*}[ht]
    \centering
    \begin{tabular}{c|cc|cc|cc}
    \toprule
        & \multicolumn{2}{c|}{50\% agreement} & \multicolumn{2}{c|}{80\% agreement} & \multicolumn{2}{c}{100\% agreement} \\
        Method & Precision@k & Recall@k & Precision@k & Recall@k & Precision@k & Recall@k \\
        \midrule
        \midrule
        Loss & 0.38 & 0.037 & 0.28 & 0.039 & 0.15 & 0.036 \\
        \midrule
        Confidence & 0.37 & 0.037 & 0.27 & 0.038 & 0.16 & 0.037 \\
        \midrule
        Parameter Gradient Norm & 0.38 & 0.038 & 0.28 & 0.038 & 0.15 & 0.035 \\
        \midrule
        Input Gradient Norm & 0.39 & 0.038 & 0.28 & 0.038 & 0.16 & 0.037 \\
        \midrule
        SHAP Values & 0.14 & 0.014 & 0.10 & 0.013 & 0.05 & 0.012 \\
        \midrule
        \textbf{LT-IQR} & \bm{\mathrm{0.96}} & \bm{\mathrm{0.094}} & \bm{\mathrm{0.90}} & \bm{\mathrm{0.125}} & \bm{\mathrm{0.77}} & \bm{\mathrm{0.180}} \\
        \bottomrule
    \end{tabular}
    \caption{Precision@k and Recall@k for vulnerability detection methods using agreement-based definition with varying agreement thresholds. Reported at fixed FPR=$10^{-3}$ and $k=1\%$ on WideResNet28-2, CIFAR-10 with varying \textbf{target model} random seeds.} 
    \label{tab:target_model_overlap}
\end{table*}

\section{DP-SGD}
\label{app:dp_sgd}

We evaluate our vulnerability detection method against models trained with Differentially Private Stochastic Gradient Descent (DP-SGD)~\cite{abadi2016deep}, which provides formal privacy guarantees by clipping per-sample gradients and adding calibrated noise.

We train WideResNet28-2 on CIFAR-10 with gradient norm clipping thresholds $C \in \{1, 3, 10\}$ and noise scale $\sigma \in \{0.0, 0.1, 1.0\}$. Following standard DP-SGD practice, we remove BatchNorm layers using the Opacus library~\cite{yousefpour2021opacus}. Due to computational expense, we use 16 shadow models for LiRA instead of our standard 256.

\begin{figure}
    \centering
    \includegraphics[width=0.45\textwidth]{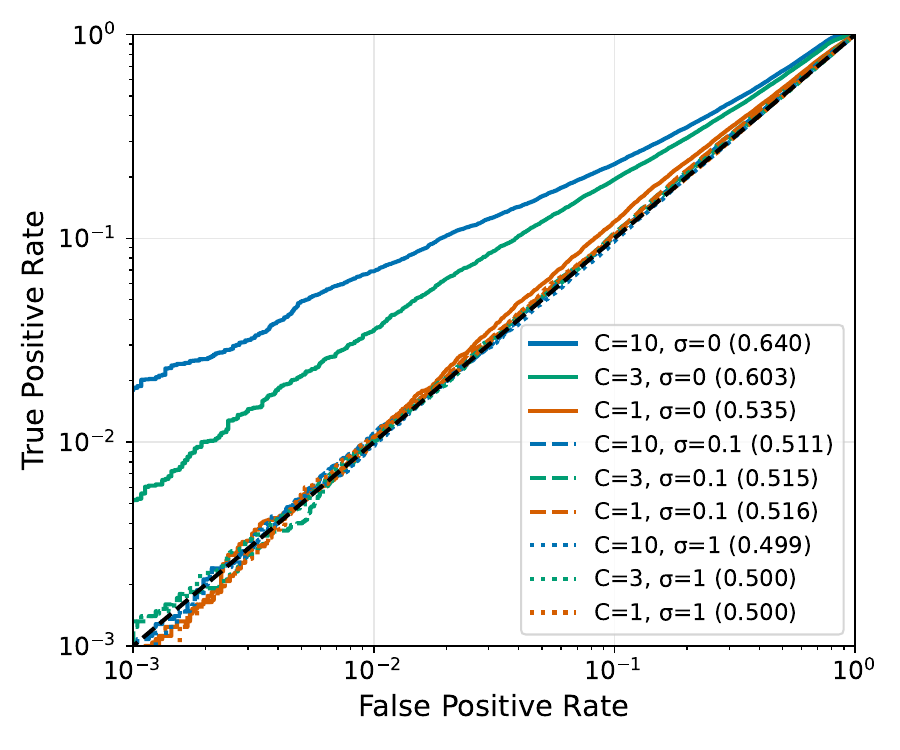}
    \caption{ROC curves for LiRA attack against DP-SGD trained models with varying clipping thresholds $C$ and noise scales $\sigma$ (AUC values in parentheses)}
    \label{fig:dp_roc}
\end{figure}

Figure~\ref{fig:dp_roc} shows LiRA ROC curves across varying noise levels. Even modest noise injection ($\sigma=0.1$) or strict gradient clipping ($C=1$) reduces LiRA performance to near-random chance, confirming that DP-SGD effectively mitigates membership inference attacks~\cite{carlini2022membership}.

For meaningful evaluation with sufficient vulnerable points, we focus on two configurations with better than random performance: $C\in\{10,3\}, \sigma=0.0$. We also adopt FPR threshold of $10^{-2}$ and $k=5\%$.

\begin{table}
    \centering
    \begin{tabular}{c|cc|cc}
    \toprule
        & \multicolumn{2}{c|}{Precision@k} & \multicolumn{2}{c}{Recall@k}\\
        Method & $C=10$ & $C=3$ & $C=10$ & $C=3$ \\
        \midrule
        \midrule
        Loss & 0.12 & 0.01 & 0.086 & 0.011 \\
        \midrule
        Confidence & 0.14 & 0.05 & 0.104 & 0.075\\
        \midrule
        \makecell{Parameter\\Gradient Norm} & 0.12 & 0.02 & 0.084 & 0.035\\
        \midrule
        \makecell{Input\\Gradient Norm} & 0.11 & 0.03 & 0.078 & 0.036\\
        \midrule
        SHAP Values & 0.13 & 0.07 & 0.095 & 0.098\\
        \midrule
        \textbf{LT-IQR} & \bm{\mathrm{0.37}} & \bm{\mathrm{0.15}} & \bm{\mathrm{0.272}} & \bm{\mathrm{0.216}}\\
        \bottomrule
    \end{tabular}
    \caption{Precision@k and Recall@k on identifying points vulnerable to LiRA attack under DP-SGD training. Reported at fixed FPR=$10^{-2}$ and $k=5\%$ with varying gradient clipping thresholds for WideResNet28-2 on CIFAR-10.} 
    \label{tab:dp_vary_c}
\end{table}

Table~\ref{tab:dp_vary_c} shows that LT-IQR consistently outperforms baseline methods across all DP-SGD configurations tested. For instance, LT-IQR achieves 0.37 precision with $C = 10$, when the best baseline method only gets 0.14. While absolute performance decreases under privacy-preserving training -- as expected given reduced vulnerable points and the impact of gradient clipping -- our method maintains its relative advantage over traditional predictors.

\section{Spearman's rank correlation}
\label{sec:spearman}

\begin{table*}
    \centering
    \begin{tabular}{c|c|c|c|c|c}
    \toprule
        & \multicolumn{5}{c}{Spearman's rank correlation} \\
         & \multicolumn{5}{c}{} \\
        & CIFAR-10 & CIFAR-10 & CIFAR-10 & CIFAR-100 &CINIC-10 \\
        Method & RN-20 & WRN40-4 & WRN28-2 & WRN28-2 & WRN28-2 \\
        \midrule
        \midrule
        Loss & 0.33 & 0.01 & 0.16 & 0.26 & 0.21  \\
        \midrule
        Confidence & 0.32 & 0.02 & 0.18 & 0.25 & 0.22 \\
        \midrule
        Parameter Gradient Norm & 0.32 & 0.08 & 0.21 & 0.26 & 0.25 \\
        \midrule
        Input Gradient Norm & 0.32 & 0.15 & 0.23 & 0.27 & 0.25 \\
        \midrule
        SHAP Values & 0.03 & 0.16 & 0.15 & 0.06 & 0.18 \\
        \midrule
        \textbf{LT-IQR} & \bm{\mathrm{0.50}} & \bm{\mathrm{0.65}} & \bm{\mathrm{0.62}} & \bm{\mathrm{0.75}} & \bm{\mathrm{0.69}} \\
        \bottomrule
    \end{tabular}
    \caption{Spearman's rank correlation between vulnerability prediction methods and LiRA attack rankings across different model architectures and datasets.} 
    \label{tab:spearman}
\end{table*}

\begin{figure}
    \centering
    \includegraphics[width=0.4\textwidth]{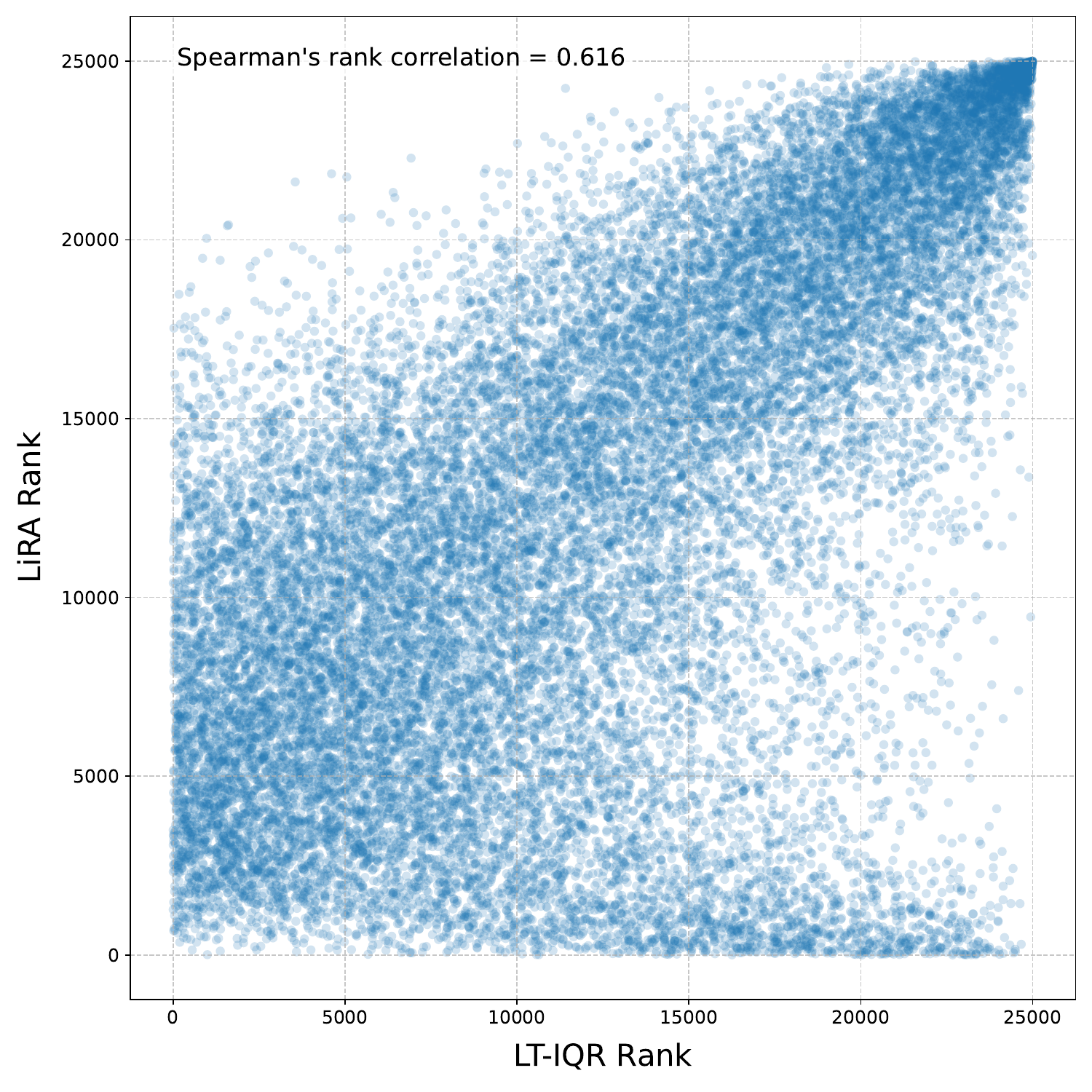}
    \caption{Scatter plot showing the correlation between LT-IQR vulnerability rankings and LiRA attack rankings for training set members on CIFAR-10 using WideResNet28-2, where higher rank indicates higher vulnerability (Spearman's rank correlation = 0.616)}
    \label{fig:spearman}
\end{figure}

Our primary evaluation focused on Precision@k and Recall@k to emphasize the identification of the most vulnerable minority of points. Prior research demonstrated that the privacy risk is distributed unevenly~\cite{carlini2022privacy, carlini2022membership}, with TPR at low FPR thresholds widely adopted as an MIA success measure instead of the AUC. For completeness, we additionally investigate Spearman's rank correlation as threshold-independent measure, which gives equal weight to correct ordering of both vulnerable and non-vulnerable points.

Figure~\ref{fig:spearman} visualizes the correlation between LT-IQR and LiRA rankings for member points in our standard setup using WideResNet28-2 on CIFAR-10. The plot reveals a dense concentration of points in the top-right corner, representing the most vulnerable samples identified by both methods. This shows that LT-IQR performs well where it matters most -- identifying the highest-risk points. We observe a general positive correlation across the full range, with a notable exception among very low-ranked points by LiRA, where LT-IQR assigns relatively high vulnerability scores. We hypothesize that these might represent challenging cases where LiRA confidently predicts members as non-members. This would suggest that both confident member and non-member predictions by LiRA may share characteristics that LT-IQR interprets as indicators of vulnerability.

Table~\ref{tab:spearman} presents Spearman's rank correlation coefficients across different model architectures and datasets. LT-IQR consistently outperforms all baseline methods across all experimental configurations, achieving correlation coefficients ranging from 0.50 to 0.75. Interestingly, we observe notable variation in the performance of simpler methods across setups. For instance, SHAP values perform particularly poorly on ResNet-20 (0.03) but achieve relatively better performance on WideResNet40-4 (0.16).

\end{document}